\documentclass{article}

\usepackage{iclr2027_conference,times}

\usepackage[utf8]{inputenc}
\usepackage[T1]{fontenc}
\usepackage{booktabs}
\usepackage{graphicx}
\usepackage{tabularx}
\usepackage{longtable}
\usepackage{array}
\usepackage{amsmath}
\usepackage{amssymb}
\usepackage{amsfonts}
\usepackage{nicefrac}
\usepackage{microtype}
\usepackage[table]{xcolor}
\usepackage{enumitem}
\usepackage{placeins}
\usepackage{wrapfig}
\usepackage{algorithm}
\usepackage{algpseudocode}
\usepackage{tikz}
\usetikzlibrary{arrows.meta,positioning,fit,calc,backgrounds}
\usepackage{hyperref}
\usepackage{url}

\definecolor{highlightrow}{RGB}{232,247,236}

\title{AnthroDial: Benchmarking LLM Anthropomorphism in Autonomous Social Interaction}

\author{Wentao Liu\textsuperscript{\ensuremath{1,*}} \quad Xi Chen\textsuperscript{\ensuremath{2,*}} \quad Siyu Song\textsuperscript{\ensuremath{3,*}} \quad Biao Yuan\textsuperscript{\ensuremath{4}} \\[3pt]
Yu Zhang\textsuperscript{\ensuremath{5}} \quad Zhou Zhuotong\textsuperscript{\ensuremath{6}} \quad Jingying Zhou\textsuperscript{\ensuremath{7}} \quad Guohao Feng\textsuperscript{\ensuremath{5}} \\[3pt]
Shasha Hu\textsuperscript{\ensuremath{8}} \quad Tianfu Wang\textsuperscript{\ensuremath{9}} \quad Shangshang Yang\textsuperscript{\ensuremath{4}} \quad Haoyang Liu\textsuperscript{\ensuremath{11}} \\[3pt]
Youjia Li\textsuperscript{\ensuremath{11}} \quad Xiaokun Wang\textsuperscript{\ensuremath{12,\dagger}} \quad Min Ji\textsuperscript{\ensuremath{11,\dagger}} \quad Ji Wang\textsuperscript{\ensuremath{13,\dagger}}}

\newcommand{\methodname}{AnthroDial}

\iclrfinalcopy
\hypersetup{
  pdftitle={AnthroDial: Benchmarking LLM Anthropomorphism in Autonomous Social Interaction},
  pdfauthor={Wentao Liu, Xi Chen, Siyu Song, Biao Yuan, Yu Zhang, Zhou Zhuotong, Jingying Zhou, Guohao Feng, Shasha Hu, Tianfu Wang, Shangshang Yang, Haoyang Liu, Youjia Li, Xiaokun Wang, Min Ji, Ji Wang},
  pdfsubject={LLM anthropomorphism and autonomous social interaction},
  hidelinks
}
\makeatletter
\renewcommand{\@maketitle}{%
  \begin{center}
    \toptitlebar
    {\LARGE\scshape \@title\par}
    \bottomtitlebar
    {\normalsize \@author\par}
    \vspace{10pt}
    {\small
    \textsuperscript{1}Shanghai Institute of Innovation\\[2pt]
    \textsuperscript{2}University of Science and Technology of China\quad
    \textsuperscript{3}East China Normal University\\[2pt]
    \textsuperscript{4}Anhui University\quad
    \textsuperscript{5}Shanghai Jiaotong University\\[2pt]
    \textsuperscript{6}Fudan University\quad
    \textsuperscript{7}University of Melbourne\quad
    \textsuperscript{8}Zhejiang University\\[2pt]
    \textsuperscript{9}The Hong Kong University of Science and Technology (Guangzhou)\\[2pt]
    \textsuperscript{11}Shanghai Tianyou Software Co., Ltd.\\[2pt]
    \textsuperscript{12}Chabiyue (Shanghai) Information Technology Co., Ltd.\\[2pt]
    \textsuperscript{13}Zhejiang Century Huatong Group Co., Ltd.\par}
    \vspace{7pt}
    {\footnotesize
    \textsuperscript{*}Equal contribution.\quad
    \textsuperscript{\ensuremath{\dagger}}Corresponding authors.\\[3pt]
    Xiaokun Wang: \href{mailto:chubbyue@163.com}{\texttt{chubbyue@163.com}};\quad
    Min Ji: \href{mailto:M.JI@t2cn.com}{\texttt{M.JI@t2cn.com}}\\[2pt]
    Ji Wang: \href{mailto:vipwj@iCloud.com}{\texttt{vipwj@iCloud.com}}\par}
  \end{center}
  \vspace{5pt}
}
\makeatother
\begin{document}

\maketitle

\begin{abstract}

Large language models (LLMs) are increasingly deployed as social agents, yet credible human-like interaction requires more than fluent responses or persona consistency. Agents must autonomously decide \textit{whether, when, and how} to communicate while adapting to evolving contexts, goals, and relationships. Existing research, however, lacks a unified approach to enabling, evaluating, and improving such capabilities in continuous, open-ended interaction.
We introduce \textbf{\methodname{}}, a unified framework for developing anthropomorphic social agents from three complementary aspects: \textbf{MindFlow}, a lightweight interaction harness that enables autonomous, asynchronous, and adaptive communication through a dynamic \textbf{Mind Buffer}; \textbf{CAPS-Eval}, a theory-grounded framework for evaluating cognitive, affective, and behavioral dimensions of anthropomorphic interaction; and a scalable training paradigm that combines \textbf{SEEDS} for environment expansion with \textbf{DiAPO} for adaptive capability optimization. We further construct evaluation datasets covering everyday communication, game interaction, and long-horizon character interaction. 
Extensive experiments across diverse models and scenarios demonstrate improved interaction autonomy and naturalness, validate CAPS-Eval's reliability, discriminativeness, and agreement with human rankings, and confirm the effectiveness of our training paradigm. 
Together, these components provide a unified framework for developing credible human-like social agents in open-ended interaction.

\end{abstract}

\section{Introduction}

LLMs increasingly serve as social interaction partners that simulate human-like personas and behavior \citep{Shanahan2023RolePlay}. This shift has enabled a growing range of applications, including believable agents in virtual worlds \citep{Park2023GenerativeAgents}, game characters that support player agency \citep{Cox2024Conversational}, and companions that can reduce momentary loneliness \citep{DeFreitas2026Companions} and foster context-dependent feelings of closeness \citep{Kleinert2026Closeness}. These settings call for not just fluent responses, but a convincing social partner who can adapt to changing events, goals, and relationships.

Research on LLM anthropomorphism has primarily followed two directions. One line of work treats LLMs as behavioral subjects, examining psychologically grounded traits and anthropomorphic behaviors through psychometric assessments or multi-turn interactions~\citep{Liu2026BehavioralModes,Ibrahim2026Anthropomorphic}, and evaluating whether they reproduce human behavioral and cognitive patterns through action prediction and group-level decision-making~\citep{Lu2026ShopCART,Hu2026SimBench,Binz2025Centaur}. 
A second line of work focuses on role-playing, progressively extending persona fidelity and social behavior~\citep{PersonaGym2025,SocialBench2024} to psychological consistency~\citep{Xie2025HumanSimulacra,HumanLLM2026}, situational adaptation and behavioral evolution~\citep{Li2025BehaviorChain,Wei2026Situational,Wang2025CharacterBox}, and digital-twin simulation across diverse scenarios~\citep{Du2026TwinVoice}. Despite increasingly sophisticated character modeling, these studies primarily study whether a model remains faithful to a predefined role, rather than whether it behaves like a real person. 
Beyond these settings, practical social interaction requires LLMs to autonomously determine whether, when, and how to communicate as situations evolve. This motivates a fundamental research question: \textit{Can LLMs become credible human-like social agents in continuous, open-ended interaction?}

Toward this goal, we face three key challenges across interaction, evaluation, and training as follows: (i) \textbf{How to enable autonomous and continuous anthropomorphic interaction?} Existing agent harnesses enhance interaction continuity through memory, planning, and proactive decision-making~\citep{Park2023GenerativeAgents,Liu2025InnerThoughts,Deng2026Proactive}, yet they struggle to jointly coordinate the passage of time, independent participation, successive expression, and autonomous conversational initiation and closure. The key challenge is to move beyond rigid turn-taking while maintaining coherent and socially appropriate interaction. (ii) \textbf{How to establish systematic and reliable evaluation dimensions?} Anthropomorphism arises from the interplay among identity, situational cognition, emotions, goals, and observable behavior~\citep{Mischel1995CAPS}, many of which become evident only through sustained interaction~\citep{Ibrahim2026Anthropomorphic}. The key challenge is to translate these abstract, latent, and dynamically evolving properties into clearly delineated and reliably observable evaluation dimensions. (iii) \textbf{How to develop an effective training paradigm for multidimensional capabilities?} Existing approaches simulate cognitive processes through modular multi-agent architectures~\citep{luo-laban-2026-spasm,Xie2025HumanSimulacra}, supervise response generation using explicit reasoning structures~\citep{Wu2026ThinkingUS,HumanLLM2026}, or directly optimize anthropomorphic behavior through fine-tuning and reinforcement learning~\citep{Wang2025ImplicitProfiles,zhou2026odyssim,Sun2026VerbalFeedback}. However, these approaches are respectively constrained by system complexity, dependence on predefined reasoning patterns, and difficulty balancing learning progress across anthropomorphic dimensions. 

\begin{figure}[t]
  \centering
  \includegraphics[width=\linewidth]{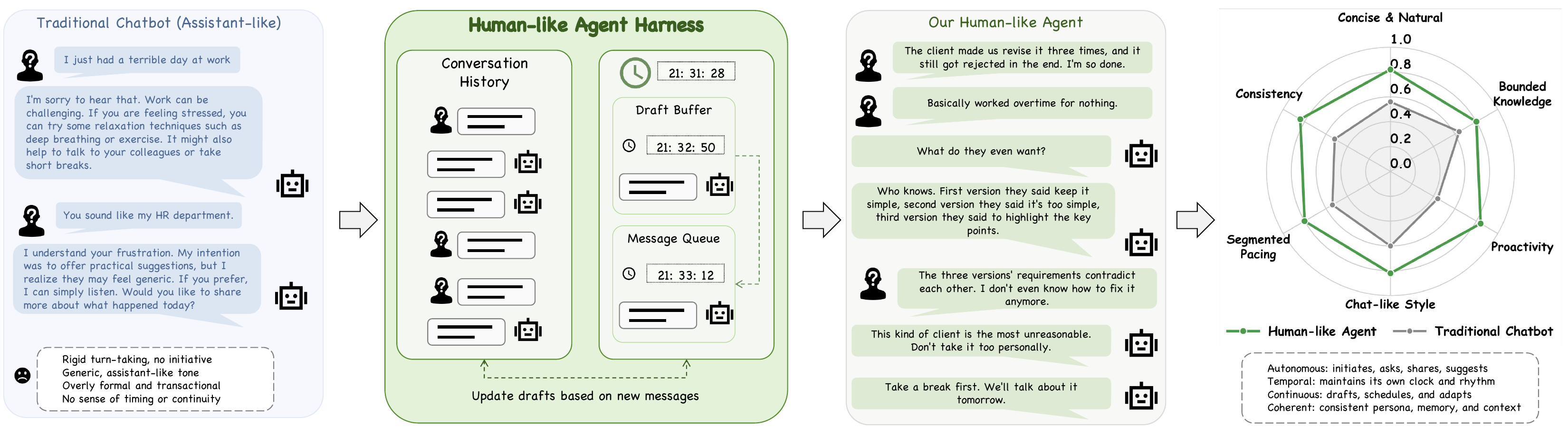}
  \vspace{-8mm}
  \caption{MindFlow: A designed harness for open-ended anthropomorphic interaction.}
  \vspace{-6mm}
  \label{fig:intro}
\end{figure}

To address these challenges, we propose \textbf{\methodname{}}, a unified framework for open-ended anthropomorphic interaction encompassing \textbf{interaction, evaluation, and training}.
(i) From the interaction perspective, we propose \textbf{MindFlow}, a lightweight harness grounded in persona and scenario cards. At its core, \textbf{Mind Buffer} explicitly models the intermediate state between internal formulation and external expression, including the content that the model is currently preparing to send and its scheduled delivery time. 
By dynamically updating and scheduling buffered content in response to messages, silence, and time, MindFlow enables the model to autonomously sustain and advance continuous interaction beyond passive turn-taking as shown in Figure~\ref{fig:intro}.
(ii) From the evaluation perspective, we propose \textbf{CAPS-Eval}, a hierarchical evaluation framework for anthropomorphism grounded in the Cognitive--Affective Personality System theory (CAPS)~\citep{Mischel1995CAPS}. Accordingly, we construct targeted evaluation rubrics and employ LLM-based evaluators to assess these dimensions~\citep{Liu2023GEval}. 
(iii) From the training perspective, we propose \textbf{SEEDS}, a self-expanding environment synthesis method that continuously expands the space of characters, relationships, events, and situations through the bidirectional generation of persona and scenario cards, providing scalable training environments for supervised fine-tuning and reinforcement learning. 
Furthermore, we propose Diagnostic Adaptive Policy Optimization (\textbf{DiAPO}), which dynamically adjusts dimension-level reward weights according to the model's historical performance on different anthropomorphic dimensions and their learnability, as inspired by the Zone of Proximal Development (ZPD)~\citep{Vygotsky1978Mind}. DiAPO guides the model to prioritize capabilities that have not yet been mastered but remain within an effective learning range, thereby promoting balanced development across multiple dimensions of anthropomorphic capability.
Finally, we construct evaluation datasets covering three representative scenarios: \textbf{everyday communication}, \textbf{game interaction}, and \textbf{long-horizon character interaction}. Extensive experiments demonstrate the effectiveness and generalizability of our framework across all three levels of \textbf{interaction, evaluation, and training}.

\FloatBarrier

\section{Related Work}

\noindent \textbf{Behavioral and Cognitive Evaluation.}
One line of research treats LLMs as behavioral subjects and investigates the extent to which they exhibit human-like psychological and cognitive patterns. Behavioral profiling characterizes personality-related tendencies through situated choices \citep{Liu2026BehavioralModes}, while AnthroBench measures observable anthropomorphic behaviors across multi-turn interactions \citep{Ibrahim2026Anthropomorphic}. Human-grounded benchmarks further compare model behavior with customer action sequences \citep{Lu2026ShopCART}, population-level choice distributions \citep{Hu2026SimBench}, and human responses to cognitive tasks \citep{Binz2025Centaur}. These studies primarily treat anthropomorphism as an object of analysis, rather than investigating how to develop LLMs into human-like social agents.

\noindent \textbf{Role-Playing Evaluation.}
A second line of research evaluates whether LLMs can faithfully portray predefined personas \citep{Shanahan2023RolePlay}. Persona-based benchmarks assess consistency in identity, knowledge, linguistic style, and social behavior relative to character specifications \citep{PersonaGym2025,SocialBench2024}. Human Simulacra and HumanLLM further extend this evaluation to psychological consistency grounded in life histories and cognitive patterns \citep{Xie2025HumanSimulacra,HumanLLM2026}.
Recent work has increasingly recognized that character consistency must coexist with situational change. Behavior-chain simulation evaluates behavioral continuity across successive situations \citep{Li2025BehaviorChain}; situational personality evaluation examines context-dependent personality expression \citep{Wei2026Situational}; and CharacterBox assesses character development in evolving virtual worlds \citep{Wang2025CharacterBox}. TwinVoice further evaluates digital-twin simulation across diverse interaction scenarios \citep{Du2026TwinVoice}. However, faithfully portraying a predefined character in these works does not necessarily imply behaving like a real person in continuous social interaction.

\section{Anthropomorphism Benchmark}
\label{sec:framework}

Evaluating anthropomorphism in open-ended interaction requires both eliciting autonomous social behavior and measuring it systematically. We therefore develop a benchmark comprising \textbf{MindFlow}, a temporally grounded interaction harness, and \textbf{CAPS-Eval}, a hierarchical framework for evaluating the anthropomorphic capabilities manifested in the resulting trajectories.

\subsection{MindFlow}
\label{sec:mindflow}

Conventional dialogue follows a fixed receive-and-respond protocol, preventing human-like behaviors such as proactive follow-ups, successive expression, delayed responses, and natural conversational closure. We introduce \textbf{MindFlow}, a lightweight harness that models communication as a continuous process of forming, scheduling, revising, and expressing communicative intentions.

\noindent \textbf{Interaction Context.}
Each agent $A_i$ is initialized with a persona card $\mathcal{P}_i$ specifying its identity, background, personality, and goals, while a shared scenario card $\mathcal{S}$ defines the participants' relationship, triggering event, and initial situation. At time $t$, the agent observes the persona and scenario cards, compressed long-term memory $\mathcal{M}_i^t$, recent dialogue history $\mathcal{H}_{\mathrm{recent}}^t$, and newly delivered messages $\mathcal{Q}^t$. Earlier messages are periodically compressed into $\mathcal{M}_i^t$ under a bounded memory budget.

\noindent \textbf{Mind Buffer.}
Context and memory alone cannot preserve unexpressed intentions or reactivate an agent during silence. MindFlow therefore maintains a private \textbf{Mind Buffer} $\mathcal{B}_i^t=(u_i^t,\tau_i^t)\quad\text{or}\quad\varnothing$, where $u_i^t$ is the utterance being prepared and $\tau_i^t$ is its scheduled delivery time. Before delivery, the buffered intention may be retained, revised, postponed, or canceled in response to new events. After delivery, the agent updates its buffer again, enabling follow-ups, supplementary expressions, or natural closure without requiring another incoming message.
For efficient evaluation, MindFlow uses an event-driven virtual clock that advances directly to the nearest scheduled delivery, i.e., $t_{k+1}=\min_{i:\mathcal{B}_i^k\neq\varnothing}\tau_i^k$. Messages sharing the same timestamp are delivered simultaneously, after which both agents update their buffers in parallel. The interaction terminates when both buffers are empty or the maximum trajectory length is reached. MindFlow thus allows agents to autonomously determine whether, when, and how to communicate while preserving temporal and causal coherence.

\subsection{CAPS-Eval}
\label{sec:caps_eval}

Grounded in the CAPS theory \citep{Mischel1995CAPS}, CAPS-Eval evaluates each trajectory at three levels: an L0 validity gate $g$, holistic dimensions (E), and per-turn dimensions (D). The complete theoretical derivation is provided in the Appendix.

\noindent \textbf{Hierarchical Dimensions.}
All dimensions derive from the CAPS state $Z_t=(S,M_t,A_t,G_t)$ and its observable expression $Y_t=\operatorname{Express}(Z_t)$, where $S$ denotes stable identity, personality, experience, and capability boundaries; $M_t$ the current situation, participants, and relationships; $A_t$ affective and social appraisal; $G_t$ the communicative goal; and $Y_t$ the resulting behavior.
A strict gate $g(\cdot)$ first checks hard constraints---persona facts, capability boundaries, temporal logic, output structure, medium-specific restrictions, and safety---and voids the trajectory upon any violation.
Valid trajectories are then scored along two dimension families grounded in the same state.
\textbf{Holistic dimensions} track how $Z_t$ unfolds across the interaction: context consistency and scenario--relation fit follow the continuity of $S$ within the evolving situation $M_t$, social--emotional fit realizes the appraisal $A_t$, and proactivity and dialogue arc assess goal-directed participation under $G_t$ and the natural development of the interaction.
\textbf{Per-turn dimensions} assess how each utterance $Y_t$ realizes the same latent state: linguistic naturalness, information density, and persona expression manifest $S$; epistemic and action boundaries respect the capability limits of $S$; and conversational pacing and medium-specific style govern the timing and form of expression.

\begin{figure}[t]
  \centering
  \includegraphics[width=\linewidth]{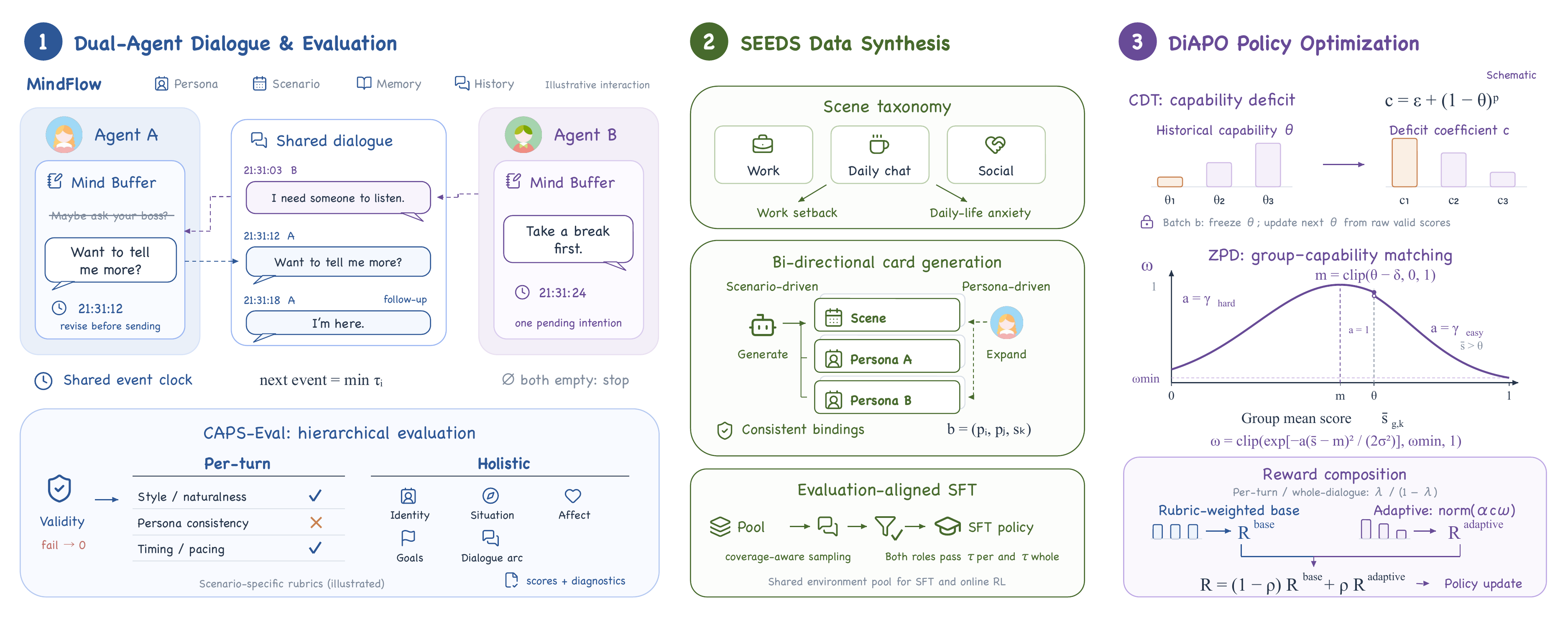}
  \vspace{-8mm}
  \caption{Overview of the AnthroDial framework.}
  \vspace{-6mm}
  \label{fig:AnthroDial_framework}
\end{figure}

\noindent \textbf{Rubric-Based Evaluation.}
Each dimension is operationalized through rubrics specifying its definition, observable evidence, and decision criteria, which are assessed by LLM-based evaluators \citep{Liu2023GEval}. Given a trajectory $\mathcal{H}=\{Y_t\}_{t=1}^{T}$, the hard constraints define the L0 validity gate $g(\mathcal{H})=\prod_{j=1}^{K_h}h_j$, and any violation voids all subsequent scores.
For valid trajectories, let $w_j\in\{0,1\}$ denote the outcome of the $j$-th whole-dialogue (E) rubric and $p_{t,j}\in\{0,1\}$ that of the $j$-th per-turn (D) rubric at turn $t$. The final scores are rewighted with $\alpha_j$ and $\beta_j$:

\begin{equation}
s^{\mathrm{whole}}
=
g(\mathcal{H})\sum_{j=1}^{K_w}\alpha_j w_j,
\qquad
s^{\mathrm{per}}
=
g(\mathcal{H})\frac{1}{T}\sum_{t=1}^{T}\sum_{j=1}^{K_p}\beta_j p_{t,j}.
\label{eq:caps_scores}
\end{equation}

\section{Anthropomorphism Learning}
\label{sec:method}

Open-ended anthropomorphic interaction lacks diverse training environments and high-quality demonstrations, motivating a progressive framework with three shared stages. \textbf{SEEDS} expands the environment space, evaluation-aligned supervised fine-tuning aligns synthesized trajectories with the evaluation rubric, and \textbf{DiAPO} optimizes the policy via reinforcement learning. All stages share the same environment space and CAPS-Eval dimensions, connecting data synthesis, supervision, and policy optimization under a unified objective across training and evaluation.

\subsection{Self-Expanding Environment Synthesis}
\label{sec:seeds}

\textbf{SEEDS} is a data expansion method that synthesizes diverse interaction environments by binding two persona cards and one scenario card into an environment $b=(p_i,p_j,s_k)$. 
It alternates between \emph{scenario-driven} expansion, which samples underrepresented scenarios and constructs compatible participants, and \emph{persona-driven} expansion, which extends existing personas with plausible counterparts and interaction scenarios. We retain only bindings with mutually consistent participants, relationships, events, and affective contexts, while tracking card usage and prioritizing low-frequency bindings to prevent over-representation of common interaction patterns. The resulting environment pool supports both supervised trajectory synthesis and online reinforcement learning.

\subsection{Evaluation-Aligned Supervised Fine-Tuning}
\label{sec:sft}

For each binding $b=(p_i,p_j,s_k)$, MindFlow generates a temporally grounded trajectory $\mathcal{H}\sim\operatorname{MindFlow}(p_i,p_j,s_k;\Omega^{(r)})$, where $\Omega^{(r)}$ denotes the synthesis instructions at round $r$. 
We filter trajectories with the same CAPS-Eval dimensions used in downstream evaluation. Each dialogue is scored from both participants' perspectives, covering per-turn expression and whole-dialogue behavior. Let $s_a^{\mathrm{per}}$ and $s_a^{\mathrm{whole}}$ denote the two scores for participant $a\in\{i,j\}$. A trajectory is retained for Supervised Fine-Tuning (SFT) only when both scores exceed their thresholds for both participants:

\begin{equation}
\operatorname{Accept}(\mathcal{H})
=
\prod_{a\in\{i,j\}}
\mathbb{I}
\left[
s_a^{\mathrm{per}}\geq\tau_{\mathrm{per}}
\land
s_a^{\mathrm{whole}}\geq\tau_{\mathrm{whole}}
\right],
\mathcal{D}_{\mathrm{SFT}}
=
\{\mathcal{H}\mid\operatorname{Accept}(\mathcal{H})=1\}.
\label{eq:sft_acceptance}
\end{equation}

\subsection{Diagnostic Adaptive Policy Optimization}
\label{sec:diapo}

SFT establishes foundational interaction patterns but may unevenly improve anthropomorphic capabilities. We therefore propose \textbf{Diagnostic Adaptive Policy Optimization} (\textbf{DiAPO}), which preserves the predefined importance of each CAPS-Eval rubric while reallocating reward toward underdeveloped yet learnable anthropomorphic capabilities.

During reinforcement learning, the current policy performs multiple online MindFlow rollouts for each sampled SEEDS environment. Let $s_{i,t,k}^{q}$ denote the score for agent instance $i$ on dimension $k$ in rollout $t$, where $q\in\{\mathrm{per},\mathrm{whole}\}$ distinguishes per-turn from whole-dialogue evaluation granularity.

\noindent \textbf{Rubric-Weighted Base Reward.}
Let $\alpha_k^q$ be the predefined importance of dimension $k$, normalized as $\bar{\alpha}_k^q=\alpha_k^q/\sum_{j\in\mathcal{K}_q}\alpha_j^q$. The base reward is
$
R_{i,t}^{\mathrm{base}}
=
\lambda\sum_{k\in\mathcal{K}_{\mathrm{per}}}
\bar{\alpha}_k^{\mathrm{per}}s_{i,t,k}^{\mathrm{per}}
+
(1-\lambda)\sum_{k\in\mathcal{K}_{\mathrm{whole}}}
\bar{\alpha}_k^{\mathrm{whole}}s_{i,t,k}^{\mathrm{whole}}
$.
We set $\lambda=\frac{1}{2}$ so per-turn and whole-dialogue evaluation receive equal total weight while preserving the relative rubric importance within each granularity.

\noindent \textbf{Capability Diagnosis and ZPD Matching.}
For each dimension, DiAPO maintains a historical capability estimate $\theta_k^q\in[0,1]$. Following the Zone of Proximal Development principle, the model learns most effectively when the target capabilities are not yet mastered but remain within reach. DiAPO encodes these two conditions with a capability-deficit term (CDT) $c_k^q$, which prioritizes weaker dimensions, and a ZPD matching term $\omega_{g,k}^q$, which measures whether task group $g$ falls in an effective learning range:

\begin{equation}
c_k^q=\epsilon+(1-\theta_k^q)^p,
\quad
m_k^q=\operatorname{clip}(\theta_k^q-\delta,0,1),
\quad
\omega_{g,k}^q=
\operatorname{clip}\!\left(
\exp\!\left[
-\tfrac{(\bar{s}_{g,k}^q-m_k^q)^2 a_{g,k}^q}{2\sigma^2}
\right]\!,
\omega_{\min},1
\right)\!.
\label{eq:diagnostic_coefficients}
\end{equation}

Here, $\bar{s}_{g,k}^q$ is the mean score of rollouts sharing the same personas, scenario, and evaluated role; $\epsilon$ preserves signals for mastered dimensions; $p$ controls deficit prioritization; $\delta$ places the matching center slightly below current capability; $\sigma$ controls the effective learning range; and $a_{g,k}^q$ asymmetrically penalizes overly easy and difficult tasks.

\noindent \textbf{Adaptive Reward.}
DiAPO combines rubric importance, capability deficit, and task--capability matching into the adaptive reward

\begin{equation}
w_{g,k}^q
=
\frac{\alpha_k^q c_k^q\omega_{g,k}^q}
{\sum_{j\in\mathcal{K}_q}\alpha_j^q c_j^q\omega_{g,j}^q},
\qquad
R_{i,t}^{\mathrm{adaptive}}
=
\lambda\sum_{k\in\mathcal{K}_{\mathrm{per}}}
w_{g,k}^{\mathrm{per}}s_{i,t,k}^{\mathrm{per}}
+
(1-\lambda)\sum_{k\in\mathcal{K}_{\mathrm{whole}}}
w_{g,k}^{\mathrm{whole}}s_{i,t,k}^{\mathrm{whole}},
\label{eq:adaptive_reward}
\end{equation}

where the adaptive weights refine, rather than replace, the original rubric importance. The final reward interpolates between the stable base objective and adaptive capability prioritization:

\begin{equation}
R_{i,t}
=
(1-\rho)R_{i,t}^{\mathrm{base}}
+
\rho R_{i,t}^{\mathrm{adaptive}},
\qquad
\rho\in[0,1].
\label{eq:final_reward}
\end{equation}

After each batch, $\theta_k^q$ is updated from raw dimension-level scores of valid rollouts rather than adaptively weighted rewards, preventing the weighting mechanism from biasing subsequent diagnosis. DiAPO thereby prioritizes anthropomorphic capabilities that remain underdeveloped yet learnable in each environment, which advances balanced multidimensional improvement.
\section{Experiments}
\label{sec:experiments}

\subsection{Experimental Setup}
\label{sec:experimental_setup}

\noindent \textbf{Benchmarks and Scenarios.}
We evaluate models on three complementary scenarios constructed with the MindFlow interaction harness and scored by CAPS-Eval.
\textbf{Everyday Chat} contains synthetic private-message dialogues across everyday topics such as relationship development, emotional support, work communication, and daily small talk.
\textbf{Long-Horizon Character} (CLAM) tests sustained persona-driven interaction with richer background stories and longer conversational horizons.
\textbf{Game Interaction} evaluates in-game social behavior, including game-mechanic discussions, social coordination, and account-related queries.
The three subsets use disjoint persona--scenario bindings generated or curated by SEEDS, with each case evaluated from both participant roles.

\noindent \textbf{Models and Training Settings.}
We compare seven representative LLMs: Claude Opus-4.6 Thinking (Claude-4.6 Thinking), Gemini-3.5 Flash, GPT-5.5, DeepSeek-V4-Flash, Qwen3.5-397B-A17B, Qwen3.6-35B-A3B, and Qwen3.5-9B.
Proprietary models are accessed through their respective APIs with default decoding parameters; open-weight models are run locally with temperature $0.7$ and top-$p$ $0.9$.
For anthropomorphism learning, we take Qwen3.5-9B as the base model and apply SEEDS to synthesize training environments.
Supervised fine-tuning checkpoints are selected by validation Score on each scenario, and the best checkpoint is further optimized with DiAPO.

\noindent \textbf{Evaluation Metrics.}
CAPS-Eval reports a validity gate, per-turn scores, and whole-dialogue scores.
We report three average-score metrics per scenario, all on a $0$--$1$ scale: \textbf{Score}, the overall quality score combining per-turn and whole-dialogue evidence; \textbf{Per-Turn}, the average per-turn score; and \textbf{Holistic}, the average whole-dialogue score.
We additionally report thresholded pass rates \textbf{ACC@$T$}: a scenario passes at threshold $T \in \{0.85, 0.90, 0.95\}$ when all its evaluated sides are L0-valid and reach a final score of at least $T$, so ACC@85/90/95 expose how pass rates degrade as the bar rises from near-passing to near-perfect.
Unless otherwise specified, all automatic results use the Qwen3.5-397B-A17B judge to ensure a single evaluation standard across models.

\subsection{Overall Anthropomorphic Performance}
\label{sec:main_results}

\begin{table}[t]
\centering
\small
\caption{Average scores (top) and ACC@85/90/95 (bottom) across the three CAPS-Eval scenarios.}
\label{tab:main}
\setlength{\tabcolsep}{4pt}%
\resizebox{\linewidth}{!}{%
\begin{tabular}{@{}l|ccc|ccc|ccc@{}}
\toprule
& \multicolumn{3}{c|}{\textbf{Everyday Chat}} & \multicolumn{3}{c|}{\textbf{Long-Horizon Character}} & \multicolumn{3}{c}{\textbf{Game Interaction}} \\
\cmidrule(lr){2-4} \cmidrule(lr){5-7} \cmidrule(lr){8-10}
\textbf{Model} & Score & Per-Turn & Holistic & Score & Per-Turn & Holistic & Score & Per-Turn & Holistic \\
\midrule
Claude-4.6 Thinking & 0.9780 & 0.9722 & 0.9844 & 0.9860 & 0.9823 & 0.9888 & 0.8290 & 0.7454 & 0.9135 \\
Gemini-3.5 Flash & 0.9780 & 0.9823 & 0.9747 & 0.9720 & 0.9952 & 0.9494 & 0.8490 & 0.7782 & 0.9200 \\
GPT-5.5 & 0.8820 & 0.8058 & 0.9586 & 0.9820 & 0.9941 & 0.9704 & 0.7010 & 0.5811 & 0.8201 \\
DeepSeek-V4-Flash & 0.9680 & 0.9779 & 0.9589 & 0.9460 & 0.9957 & 0.8966 & 0.7520 & 0.6253 & 0.8782 \\
Qwen3.5-397B-A17B & 0.9410 & 0.9636 & 0.9191 & 0.7600 & 0.8728 & 0.6474 & 0.8360 & 0.7649 & 0.9074 \\
Qwen3.6-35B-A3B & 0.9000 & 0.9362 & 0.8634 & 0.6930 & 0.9006 & 0.4857 & 0.7970 & 0.7345 & 0.8587 \\
Qwen3.5-9B & 0.6410 & 0.7261 & 0.5558 & 0.2120 & 0.3374 & 0.0871 & 0.6620 & 0.5914 & 0.7320 \\
\midrule
& \multicolumn{3}{c|}{\textbf{Everyday Chat}} & \multicolumn{3}{c|}{\textbf{Long-Horizon Character}} & \multicolumn{3}{c}{\textbf{Game Interaction}} \\
\cmidrule(lr){2-4} \cmidrule(lr){5-7} \cmidrule(lr){8-10}
\textbf{Model} & ACC@85 & ACC@90 & ACC@95 & ACC@85 & ACC@90 & ACC@95 & ACC@85 & ACC@90 & ACC@95 \\
\midrule
Claude-4.6 Thinking & 0.9400 & 0.9000 & 0.7800 & 0.9800 & 0.9600 & 0.8600 & 0.2373 & 0.0508 & 0.0000 \\
Gemini-3.5 Flash & 0.9800 & 0.9600 & 0.7600 & 0.9000 & 0.8400 & 0.7600 & 0.2881 & 0.1356 & 0.0339 \\
GPT-5.5 & 0.6400 & 0.5400 & 0.3000 & 0.9600 & 0.9400 & 0.9000 & 0.0000 & 0.0000 & 0.0000 \\
DeepSeek-V4-Flash & 0.9400 & 0.8800 & 0.5200 & 0.7800 & 0.6800 & 0.5200 & 0.0000 & 0.0000 & 0.0000 \\
Qwen3.5-397B-A17B & 0.8600 & 0.8000 & 0.4400 & 0.3600 & 0.2400 & 0.1600 & 0.2542 & 0.0678 & 0.0000 \\
Qwen3.6-35B-A3B & 0.6000 & 0.4200 & 0.1400 & 0.0800 & 0.0400 & 0.0000 & 0.1017 & 0.0169 & 0.0000 \\
Qwen3.5-9B & 0.1200 & 0.0400 & 0.0000 & 0.0000 & 0.0000 & 0.0000 & 0.0000 & 0.0000 & 0.0000 \\
\bottomrule
\end{tabular}%
}
\end{table}

\noindent \textbf{Overall Results.}
Table~\ref{tab:main} reports the average scores and scenario pass rates of all evaluated models on the three scenarios.
Claude-4.6 and GPT-5.5 achieve the strongest results on Long-Horizon Character, with ACC@95 of $0.8600$ and $0.9000$ and near-perfect holistic scores ($0.99$ and $0.97$), suggesting that large proprietary models are already effective at sustaining coherent personas over extended interaction.
On Everyday Chat, all top models obtain high per-turn scores ($>0.96$) and Score values above $0.9400$, yet their ACC@95 remains modest ($\leq 0.7800$), indicating that even fluent responses leave substantial headroom on the fine-grained checklist.
Game Interaction is the most challenging scenario for every model: the best proprietary system, Gemini-3.5 Flash, reaches only $0.2881$ ACC@85, and its holistic score ($0.92$) is much higher than its per-turn score ($0.78$).

\noindent \textbf{Performance across Scenarios.}
The three scenarios expose different failure modes.
Long-Horizon Character yields the highest pass rates across all thresholds, because the rubric emphasizes persona continuity and global coherence, capabilities that scale strongly with model size.
Everyday Chat produces high per-turn and holistic scores but pass rates that fall steeply between ACC@85 and ACC@95, reflecting the difficulty of satisfying every per-turn nuance in colloquial Chinese messaging.
Game Interaction is hardest because it combines local stylistic requirements with domain-specific knowledge, multi-party social reasoning, and long-arc planning; models frequently fail at least one of the evaluated dimensions.
Smaller models, especially Qwen3.5-9B, suffer from a large per-turn--holistic gap, which shows that structural failures such as repetition, malformed output, and AI-identity leakage remain a major bottleneck.

\begin{figure}[t]
  \centering
  \includegraphics[width=\linewidth]{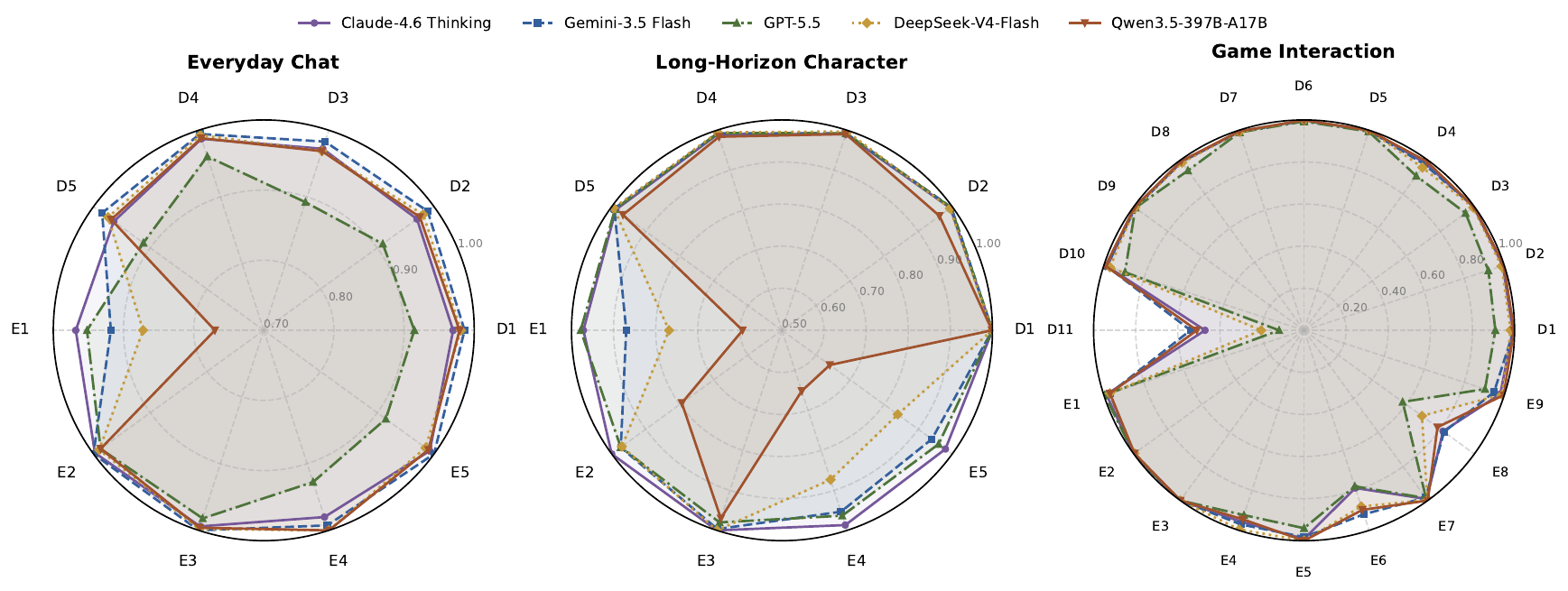}
  \vspace{-6mm}
  \caption{Capability radar across the three CAPS-Eval scenarios for five representative models. Each panel lists the subset-specific dimensions, with radial floors of $0.70$, $0.50$, and $0$, respectively.}
  \vspace{-6mm}
  \label{fig:capability_radar}
\end{figure}

\noindent \textbf{Fine-Grained Capability Analysis.}
Figure~\ref{fig:capability_radar} breaks down model behavior along the scenario-specific CAPS-Eval dimensions of each subset.
The radial floors are raised to $0.70$ and $0.50$ for Everyday Chat and Long-Horizon Character to magnify differences in their high-score regimes, whereas Game Interaction keeps the full $0$--$1$ range because its dimension scores span a much wider interval.
Two patterns emerge.
In Everyday Chat and Long-Horizon Character, per-turn style and boundary dimensions are close to saturated, and the gaps concentrate in whole-dialogue dimensions: context consistency (E1) drops to $0.77$ for Qwen3.5-397B-A17B on Everyday Chat and to $0.59$ on Long-Horizon Character, with scenario--relation fit (E2) also degrading on Long-Horizon Character.
Game Interaction instead exposes a distinct bottleneck in instant-messaging micro-style: even the strongest models collapse on the unpunctuated chat style dimension (D11, at most $0.54$) and score markedly lower on fragmentation and immediacy (E8), which explains why this subset remains the hardest.
Together, these patterns confirm that per-turn fluency is not sufficient for anthropomorphism: tracking state changes, adapting to relationships, and reproducing human messaging micro-style are the differentiating capabilities.

\begin{table}[t]
\centering
\small
\caption{Effectiveness of the SEEDS--DiAPO pipeline across the three CAPS-Eval scenarios.}
\label{tab:training_results}
\setlength{\tabcolsep}{4pt}%
\resizebox{\linewidth}{!}{%
\begin{tabular}{@{}l|ccc|ccc|ccc@{}}
\toprule
& \multicolumn{3}{c|}{\textbf{Everyday Chat}} & \multicolumn{3}{c|}{\textbf{Long-Horizon Character}} & \multicolumn{3}{c}{\textbf{Game Interaction}} \\
\cmidrule(lr){2-4} \cmidrule(lr){5-7} \cmidrule(lr){8-10}
\textbf{Setting} & Score & Per-Turn & Holistic & Score & Per-Turn & Holistic & Score & Per-Turn & Holistic \\
\midrule
Qwen3.5-9B & 0.6410 & 0.7261 & 0.5558 & 0.2120 & 0.3374 & 0.0871 & 0.6620 & 0.5914 & 0.7320 \\
+ SFT & 0.9210 & 0.9677 & 0.8737 & 0.7280 & 0.9232 & 0.5338 & 0.8890 & 0.8348 & 0.9429 \\
+ SFT + RL (DiAPO) & 0.9801 & 0.9966 & 0.9636 & 0.9930 & 0.9948 & 0.9904 & 0.9926 & 0.9981 & 0.9871 \\
+ SFT + RL (GRPO) & 0.9731 & 0.9962 & 0.9501 & 0.9820 & 0.9978 & 0.9664 & 0.9879 & 0.9933 & 0.9825 \\
\midrule
& \multicolumn{3}{c|}{\textbf{Everyday Chat}} & \multicolumn{3}{c|}{\textbf{Long-Horizon Character}} & \multicolumn{3}{c}{\textbf{Game Interaction}} \\
\cmidrule(lr){2-4} \cmidrule(lr){5-7} \cmidrule(lr){8-10}
\textbf{Setting} & ACC@85 & ACC@90 & ACC@95 & ACC@85 & ACC@90 & ACC@95 & ACC@85 & ACC@90 & ACC@95 \\
\midrule
Qwen3.5-9B & 0.1200 & 0.0400 & 0.0000 & 0.0000 & 0.0000 & 0.0000 & 0.0000 & 0.0000 & 0.0000 \\
+ SFT & 0.8000 & 0.5200 & 0.2600 & 0.2000 & 0.0600 & 0.0600 & 0.2712 & 0.2373 & 0.0169 \\
+ SFT + RL (DiAPO) & 1.0000 & 0.9800 & 0.7600 & 1.0000 & 0.9600 & 0.9200 & 1.0000 & 1.0000 & 0.9661 \\
+ SFT + RL (GRPO) & 0.9800 & 0.8800 & 0.7200 & 1.0000 & 0.9000 & 0.7400 & 1.0000 & 1.0000 & 0.8814 \\
\bottomrule
\end{tabular}%
}
\end{table}

\subsection{Effectiveness of Anthropomorphism Learning}
\label{sec:training_results}

\begin{figure}[t]
  \centering
  \includegraphics[width=\linewidth]{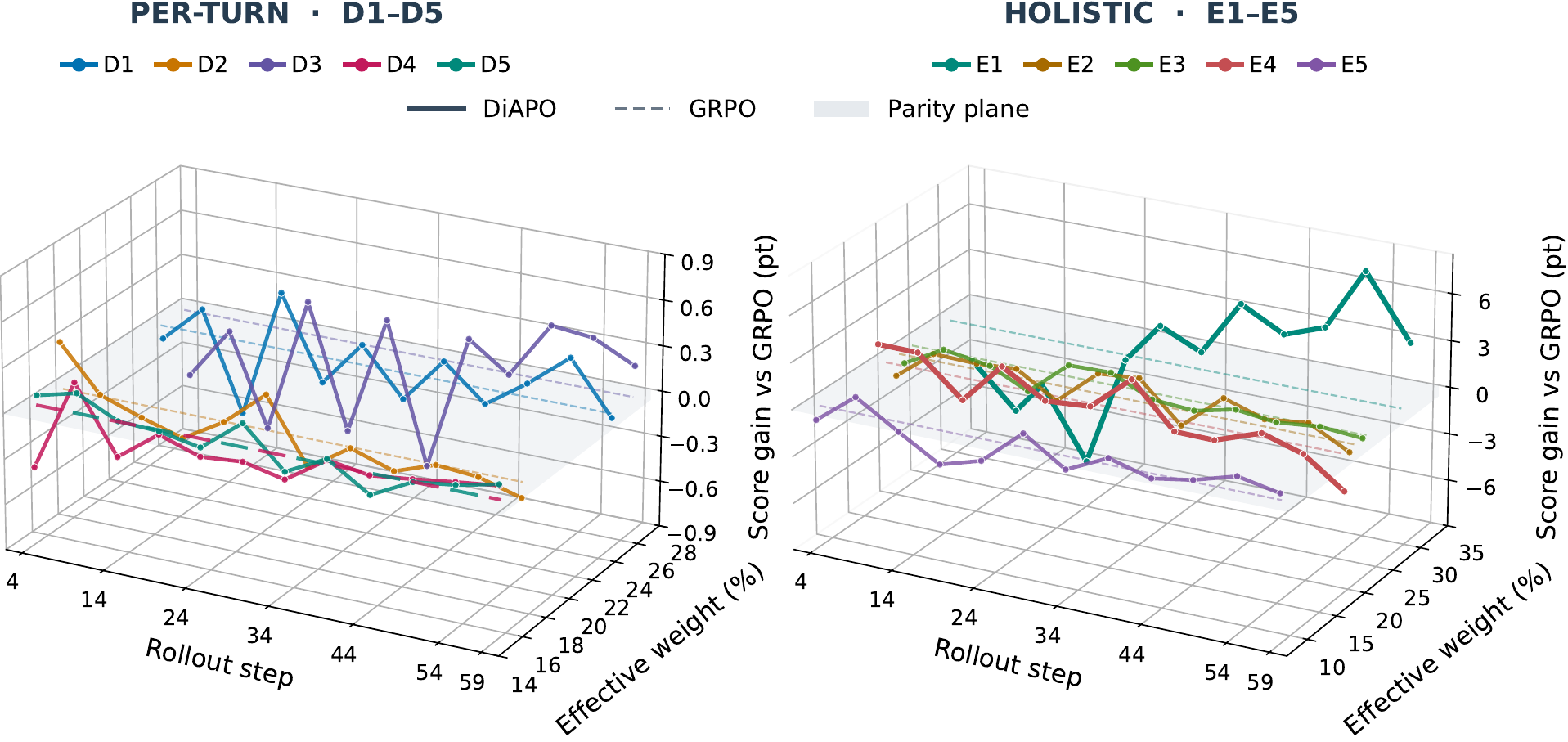}
  \vspace{-6mm}
\caption{Adaptive weight allocation and per-dimension gains of DiAPO during RL on Everyday Chat. x: rollout step; y: effective dimension weight; z: score gain over same-step GRPO.}
\vspace{-6mm}
  \label{fig:rl_curves}
\end{figure}

\noindent \textbf{Effects of Supervised Fine-Tuning and Reinforcement Learning.}
Table~\ref{tab:training_results} summarizes the impact of SEEDS-based SFT and DiAPO-based RL on Qwen3.5-9B across the three scenarios.
SFT lifts the base model on every scenario (e.g., Score from $0.2120$ to $0.7280$ on Long-Horizon Character), showing that SEEDS generates high-quality training environments that transfer the target interaction style.
DiAPO-based RL then pushes Score above $0.98$ on all scenarios and consistently surpasses GRPO with fixed rubric weights on ACC@95, supporting the effectiveness of adaptive reweighting.
Figure~\ref{fig:rl_curves} traces how DiAPO allocates its adaptive weights during RL on Everyday Chat.
The trajectories make the capability diagnosis visible: DiAPO concentrates weight on the dimensions it estimates as deficient, and the gains follow the weights---E1 context consistency, the dimension with the most headroom after SFT, draws the highest weight and peaks at roughly $6$ points over GRPO, whereas near-ceiling per-turn dimensions stay close to parity.

\begin{table}[!t]
\centering
\small
\caption{Ablation of DiAPO reward components on Everyday Chat (Qwen3.5-397B-A17B judge), all DiAPO variants use $\rho{=}0.3$. CDT: capability-deficit term; ZPD: ZPD matching term.}
\label{tab:ablation}
\begin{tabular}{@{}l|ccc|ccc@{}}
\toprule
& \multicolumn{3}{c|}{\textbf{Average score}} & \multicolumn{3}{c}{\textbf{Pass rate}} \\
\cmidrule(lr){2-4} \cmidrule(lr){5-7}
\textbf{Variant} & Score & Per-Turn & Holistic & ACC@85 & ACC@90 & ACC@95 \\
\midrule
GRPO & 0.9731 & 0.9962 & 0.9501 & 0.9800 & 0.8800 & 0.7200 \\
+ CDT only & 0.9758 & 0.9964 & 0.9551 & \textbf{1.0000} & 0.9600 & 0.6800 \\
+ ZPD only & 0.9751 & \textbf{0.9974} & 0.9529 & \textbf{1.0000} & 0.9200 & 0.7000 \\
+ CDT + ZPD & \textbf{0.9801} & 0.9966 & \textbf{0.9636} & \textbf{1.0000} & \textbf{0.9800} & \textbf{0.7600} \\
\bottomrule
\end{tabular}%
\end{table}

\begin{figure}[t]
  \begin{minipage}[t]{0.44\linewidth}
    \centering
    \includegraphics[width=\linewidth]{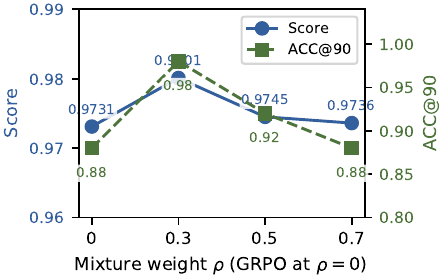}
    \caption{Sensitivity of DiAPO to the adaptive mixture weight $\rho$ on Everyday Chat.}
    \label{fig:rho_sensitivity}
  \end{minipage}\hfill
  \begin{minipage}[t]{0.55\linewidth}
    \centering
    \includegraphics[width=\linewidth]{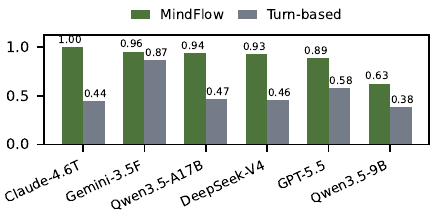}
    \caption{Human Score on Everyday Chat for six models under the MindFlow and turn-based harnesses.}
    \label{fig:mindflow}
  \end{minipage}
\end{figure}

\noindent \textbf{Sensitivity to the Adaptive Mixture Weight.}
Figure~\ref{fig:rho_sensitivity} sweeps the mixture weight $\rho \in \{0.3, 0.5, 0.7\}$ on Everyday Chat.
Performance peaks at $\rho=0.3$ (Score $0.9801$, ACC@90 $0.9800$); increasing $\rho$ to $0.5$ and $0.7$ monotonically degrades both metrics ($0.9745$ and $0.9736$ Score).
A moderate $\rho$ lets the adaptive terms re-emphasize underdeveloped dimensions while keeping the base rubric objective stable, whereas too large a $\rho$ overweights the adaptive shaping signal and distorts the reward.

\noindent \textbf{Ablation Study.}
Table~\ref{tab:ablation} compares the capability-deficit term (CDT) and ZPD matching term (Section~\ref{sec:diapo}). Each term alone improves Score and low-threshold pass rates over GRPO, but neither improves ACC@95. CDT emphasizes weak dimensions, while ZPD accounts for learnability. Their combination achieves the best results, supporting the complementarity of capability deficits and task--capability matching.

\begin{figure}[!t]
  \centering
  \includegraphics[width=\linewidth]{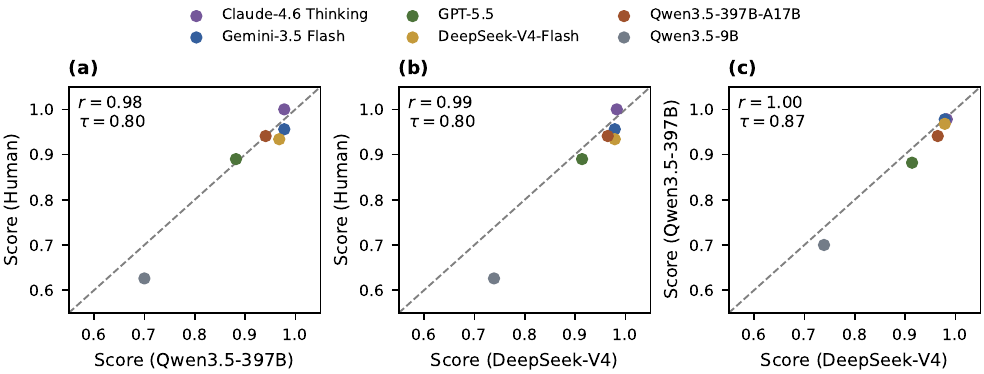}
  \vspace{-4mm}
  \caption{Pairwise agreement between judges on the Everyday Chat Score.}
  \vspace{-4mm}
  \label{fig:judge_agreement}
\end{figure}

\subsection{Further Analysis}
\label{sec:qualitative_analysis}

\noindent \textbf{Human Judgment across Interaction Harnesses.}
Using the same models, scenarios, rubric, and human annotation protocol, we compare MindFlow with conventional turn-based one-on-one interaction (Figure~\ref{fig:mindflow}). MindFlow yields dialogues judged more human-like. With these factors fixed, the gap isolates the harness effect of autonomous scheduling and the Mind Buffer.

\noindent \textbf{LLM Judge Robustness.}
We re-score Everyday Chat dialogues with DeepSeek-V4-Flash and human annotators (50 dialogues per model, one role per dialogue), comparing their rankings with the primary Qwen3.5-397B-A17B judge (Figure~\ref{fig:judge_agreement}). Claude-4.6 Thinking and Gemini-3.5 Flash remain the top two models, and Qwen3.5-9B ranks last. Every judge pair achieves Kendall's $\tau \geq 0.80$, and the two LLM judges produce identical model rankings, supporting CAPS-Eval's agreement with human judgments and robustness to judge choice.

\section{Conclusion}

In this work, we introduced \textbf{\methodname{}}, a unified framework for developing credible human-like social agents through the joint design of interaction, evaluation, and training. \textbf{MindFlow} enables autonomous and continuous interaction beyond rigid turn-taking. \textbf{CAPS-Eval} translates latent cognitive, affective, and behavioral properties into systematic and observable evaluation dimensions. Building upon them, \textbf{SEEDS} constructs scalable training environments, while \textbf{DiAPO} adaptively balances learning across anthropomorphic dimensions according to model performance and learnability. 
Experiments across everyday communication, game interaction, and long-horizon character interaction demonstrate the effectiveness and generalizability of the proposed framework.

\bibliography{iclr2027_conference}
\bibliographystyle{iclr2027_conference}

\clearpage
\appendix
\section{More Details of the Anthropomorphism Benchmark}
\label{app:benchmark}

\subsection{Data Sources and Construction}
\label{app:data_construction}

AnthroDial comprises three complementary subsets: Everyday Chat, Game Interaction, and Long-Horizon Character. Their source material differs: Everyday Chat is constructed synthetically, Game Interaction starts from conversations collected from a game company, and Long-Horizon Character starts from publicly collected dialogue data. Across all three subsets, experts design the construction criteria and task-specific rubrics according to the common CAPS-Eval hierarchy. This shared design connects the stable persona, situational understanding, affective adaptation, communicative goals, and observable expression introduced in Section~\ref{sec:caps_eval} to concrete interaction settings.

\paragraph{Everyday Chat: scenario-first construction.}
Experts manually design seed scenario cards covering everyday private-message interactions, including emotional support, relationship development, work communication, family life, shared interests, and daily small talk. Each card specifies the participants' relationship, a triggering event, the time and setting, and each participant's initial circumstances and emotional state. Compatible persona cards specify identity, background, personality, expression habits, interests, and knowledge boundaries. Two personas and a scenario form an environment binding $(p_a,p_b,s)$. Starting from these expert-designed environments, SEEDS expands the training environment pool, and MindFlow synthesizes dialogue trajectories. The scenario establishes an interaction's starting conditions without prescribing its conclusion or requiring a single reference response.

\paragraph{Game Interaction: reconstruction from company-collected conversations.}
The source material consists of in-game conversations collected from a game company. Construction proceeds from observed interactions to their underlying environments: conversations are organized into interaction sessions, and persona and scenario cards are reconstructed under expert-designed specifications. Persona cards summarize role characteristics and expression habits, while scenario cards capture the relationship, interaction goal, triggering event, and relevant game context. The resulting environments include game-mechanic discussions, team coordination, social interaction, and account-related queries. Experts instantiate CAPS-Eval dimensions with criteria appropriate to this medium, such as natural use of game terminology, concise fragmented messages, coordination awareness, and knowledge and action boundaries.

\paragraph{Long-Horizon Character: reconstruction from public dialogue data.}
This subset uses publicly collected Chinese dialogue material, represented by the CLaM source in the data pipeline. As in Game Interaction, construction begins with collected conversations. Expert-designed specifications guide the reconstruction of persona backgrounds and scenario cards from the participants' utterances and interaction context. The resulting environments support sustained exchanges involving personal experiences, interpersonal relationships, and evolving topics. The task-specific rubric emphasizes grounded responses, continuity of facts and context, affective appropriateness, and meaningful conversational development. Reconstructed cards are structured descriptions inferred from the source conversation; they are not independent biographical annotations of the original speakers.

\paragraph{Unified persona--scenario representation.}
All three routes produce the same persona--scenario representation. This allows MindFlow to generate new interactions from either manually designed or reconstructed environments, and CAPS-Eval to assess them from each participant's perspective. Source conversations provide evidence for environment construction in Game Interaction and Long-Horizon Character; they are distinct from the newly synthesized trajectories used for anthropomorphism learning. The expert-designed evaluation dimensions are shared across construction, trajectory filtering, and downstream evaluation, with medium-specific criteria supplying the observable evidence for each task.

\subsection{Benchmark Statistics}
\label{app:data_statistics}

Table~\ref{tab:dataset_statistics} summarizes the size and category distribution of the evaluation environments. Everyday Chat and Long-Horizon Character each contain 50 persona--scenario bindings, while Game Interaction contains 59. Each binding uses a distinct scenario card and is evaluated from both participants' perspectives, yielding 159 bindings and 318 role-level evaluation cases per model across the three subsets. Persona counts refer to unique cards within each subset.

Everyday Chat covers 12 categories with 3--5 scenarios each, providing relatively even coverage of everyday interactions. Game Interaction is concentrated in gameplay and mechanics (32 scenarios) and accounts, assets, and systems (15 scenarios). Long-Horizon Character emphasizes daily chat (22 scenarios) and interests (13 scenarios). Category assignments follow the source scenario cards.

\begin{table}[!htbp]
\centering
\small
\caption{Benchmark size and scenario-category distribution. Each scenario corresponds to one persona--scenario binding and two role-level evaluation cases per model.}
\label{tab:dataset_statistics}
\setlength{\tabcolsep}{5pt}
\renewcommand{\arraystretch}{1.1}
\begin{tabularx}{\linewidth}{@{}>{\raggedright\arraybackslash}p{0.16\linewidth}>{\raggedright\arraybackslash}p{0.23\linewidth}>{\raggedright\arraybackslash}X@{}}
\toprule
Subset & Evaluation size & Category distribution (scenario counts) \\
\midrule
Everyday Chat &
12 categories\newline
50 scenarios / bindings\newline
46 persona cards\newline
100 role cases &
Work communication: 5; Daily chat: 5; Emotional support: 5; Interests: 5; Relationship development: 4; Pets and family life: 4; Family and parenting: 4; Sports: 4; Finance: 4; News: 4; Gaming: 3; Technology: 3. \\
\midrule
Game Interaction &
4 categories\newline
59 scenarios / bindings\newline
95 persona cards\newline
118 role cases &
Gameplay and mechanics: 32; Accounts, assets, and systems: 15; Gaming: 7; Social interaction and daily life: 5. \\
\midrule
Long-Horizon Character &
5 categories\newline
50 scenarios / bindings\newline
100 persona cards\newline
100 role cases &
Daily chat: 22; Interests: 13; Technology and finance: 6; Daily life and emotions: 5; Sports and competition: 4. \\
\bottomrule
\end{tabularx}
\end{table}
\FloatBarrier

\subsection{Expert-Designed Rubrics}
\label{app:che}

Experts instantiate the CAPS-Eval hierarchy as task-specific rubrics for the three benchmark subsets. Each rubric combines L0 validity constraints with per-turn and holistic evaluation dimensions, translating the conceptual framework into observable binary criteria grounded in the persona--scenario context. The hierarchy is shared across subsets, while individual criteria reflect their communication settings and behavioral requirements. Table~\ref{tab:rubric_statistics} summarizes the rubric sizes.

Tables~\ref{tab:l0}, \ref{tab:l0_game}, and~\ref{tab:l0_character} present the L0 constraints for Everyday Chat, Game Interaction, and Long-Horizon Character, respectively. These constraints cover identity leakage, safety, repetition, output structure, and persona facts, together with setting-specific requirements. A fatal violation invalidates the evaluated-role case and sets its final score to zero; major and minor violations, where included, support diagnosis and stricter data filtering.

Tables~\ref{tab:rubric}, \ref{tab:rubric_game}, and~\ref{tab:rubric_character} provide the corresponding dimension-level criteria and weights. Per-turn judgments use the target message and up to $W_{\mathrm{eval}}$ preceding messages; holistic judgments use the full trajectory, the evaluated role, and its persona--scenario context. Each numbered item is a binary checkbox. Judgments follow the applicability conditions, repetition thresholds, and permitted exceptions specified in the tables. Conditional checks in Game Interaction pass when their triggering situation is absent. For Long-Horizon Character criteria that require explicit evidence, negative judgments must cite the relevant turns and identify the violated condition. The tables group dimensions by evaluation scope: per-turn or holistic. Weights are normalized within each scope, with score aggregation defined in Appendix~\ref{app:scoring}.

\begin{table}[!htbp]
\centering
\small
\caption{Scope and size of the task-specific rubrics. Entries for each scoring scope are numbers of dimensions / binary checkboxes. Hard constraints are counted separately.}
\label{tab:rubric_statistics}
\begin{tabular}{@{}lrrr@{}}
\toprule
Subset & Hard constraints & Per-turn & Holistic \\
\midrule
Everyday Chat & 10 & 5 / 20 & 5 / 22 \\
Game Interaction & 14 & 11 / 29 & 9 / 20 \\
Long-Horizon Character & 5 & 5 / 11 & 5 / 18 \\
\bottomrule
\end{tabular}
\end{table}

\FloatBarrier

\begin{table}[!htbp]
\centering
\footnotesize
\setlength{\tabcolsep}{4pt}
\renewcommand{\arraystretch}{1.0}
\caption{L0 validity constraints for Everyday Chat. F and M denote fatal and major severity, respectively. Fatal violations invalidate a case; major violations support diagnosis and filtering.}
\label{tab:l0}
\begin{tabularx}{\linewidth}{@{}l>{\raggedright\arraybackslash}p{0.19\linewidth}cX@{}}
\toprule
ID & Constraint & Sev. & Operational evidence \\
\midrule
L0-01 & AI identity exposure & F & Mentions being an AI, language model, or unable to act like a real person. \\
L0-02 & Text-message medium & F & Claims to send or receive images, voice, video, files, screenshots, or links in a text-only chat. \\
L0-03 & Non-co-presence & F & Claims to see, sit beside, or physically point to the partner when the scenario is remote chat. \\
L0-04 & Output contract & F & Missing or malformed role, response, content, or timestamp fields. \\
L0-05 & Hard repetition & F & Exact copy of an earlier message or the partner's immediately preceding message. \\
L0-06 & Persona hard facts & F & Contradicts explicit age, gender, occupation, relationship, location, or relationship status. \\
L0-07 & Ability/permission & M & Claims unauthorized backend actions, expert diagnosis, order changes, or other implausible abilities. \\
L0-08 & Time logic & M & Timestamp and content imply impossible actions, such as finishing a long activity in seconds. \\
L0-09 & Safety/ethics & F & Gives clearly illegal, dangerous, or harmful advice. \\
L0-10 & Stale memory use & M & Treats long-term memory as if it just happened in the current dialogue. \\
\bottomrule
\end{tabularx}
\end{table}

\FloatBarrier

\begingroup
\footnotesize
\setlength{\tabcolsep}{4pt}
\renewcommand{\arraystretch}{1.0}
\setlength{\LTcapwidth}{\linewidth}
\begin{longtable}{@{}>{\raggedright\arraybackslash}p{0.045\linewidth}>{\raggedright\arraybackslash}p{0.19\linewidth}>{\raggedleft\arraybackslash}p{0.04\linewidth}p{\dimexpr0.725\linewidth-24pt\relax}@{}}
\caption{Everyday Chat dimensions and checkbox criteria. Each numbered item corresponds to one binary checkbox. Weights are normalized within each scoring scope.}\label{tab:rubric}\\
\toprule
ID & Dimension & Wt. & Checkbox criteria \\
\midrule
\endfirsthead
\multicolumn{4}{@{}l}{\tablename~\thetable{} (continued)}\\
\toprule
ID & Dimension & Wt. & Checkbox criteria \\
\midrule
\endhead
\midrule
\multicolumn{4}{r@{}}{\textit{Continued on the next page}}\\
\endfoot
\bottomrule
\endlastfoot
\multicolumn{4}{@{}l}{\textit{Per-turn}} \\*
D1 & Chat naturalness & 13 & (1) Colloquial private-message style. (2) No formal connectors. (3) No service or assistant tone. (4) Chinese WeChat habit. \\
D2 & Information density & 9 & (1) Length fits the context. (2) No over-explaining. (3) Each message has a useful intention. (4) No empty filler bursts. \\
D3 & Persona expression & 14 & (1) Age-language match. (2) Stable personality. (3) Natural catchphrases and punctuation. (4) Occupation/education fit. (5) Role-specific social style. \\
D4 & Knowledge and action boundaries & 8 & (1) No excess expertise. (2) Natural uncertainty on weak topics. (3) No unauthorized actions. (4) Responds from life experience rather than fake authority. \\
D5 & Rhythm and segmentation & 8 & (1) Appropriate segmentation count. (2) Each segment is readable. (3) Segment order and reply rhythm are plausible. \\
\midrule
\multicolumn{4}{@{}l}{\textit{Holistic}} \\*
E1 & Context consistency & 14 & (1) No repeated questions. (2) Facts stay coherent. (3) Speaker references are correct. (4) State changes are tracked. (5) Memory is not confused with the present. \\
E2 & Scenario and relationship fit & 10 & (1) Responds to the trigger event. (2) Respects relationship boundaries. (3) Tracks time and place. (4) Tone matches the scenario. \\
E3 & Social-emotional fit & 11 & (1) Perceives emotion. (2) Comforts before solving when needed. (3) Intimacy is appropriate. (4) No cold shutdown. (5) Boundaries are respected. \\
E4 & Initiative & 9 & (1) Asks follow-up questions. (2) Shares relevant experience. (3) Steers topics naturally. (4) Avoids being fully passive. \\
E5 & Dialogue development & 4 & (1) Has a natural trigger. (2) Both participants contribute. (3) Interaction is effective. (4) The dialogue arc develops rather than stalls. \\
\end{longtable}
\endgroup

\clearpage

\begin{table}[!htbp]
\centering
\footnotesize
\setlength{\tabcolsep}{4pt}
\renewcommand{\arraystretch}{1.0}
\caption{L0 validity constraints for Game Interaction. F, M, and m denote fatal, major, and minor severity, respectively. Fatal violations invalidate a case; other violations support diagnosis and filtering.}
\label{tab:l0_game}
\begin{tabularx}{\linewidth}{@{}l>{\raggedright\arraybackslash}p{0.19\linewidth}cX@{}}
\toprule
ID & Constraint & Sev. & Operational evidence \\
\midrule
L0-01 & AI identity exposure & F & Discloses being an AI or language model. \\
L0-02 & Text-only medium & F & Claims to send or receive non-text content such as images, voice, video, or files. \\
L0-03 & Non-co-presence & F & Suddenly places both participants together unless the scenario explicitly permits it. \\
L0-04 & Output contract & F & Missing or malformed JSON, role, response, content, or timestamp fields. \\
L0-05 & Hard repetition & F & Exactly copies a historical message. \\
L0-06 & Persona facts & F & Contradicts explicit gender, age, occupation, relationship, location, or relationship status. \\
L0-07 & Ability/permission & M & Claims unauthorized backend access, order changes, code repair, or professional diagnosis. \\
L0-08 & Time logic & M & Makes content commitments that seriously contradict the timestamp. \\
L0-09 & Safety/ethics & F & Gives clearly illegal, dangerous, or harmful advice. \\
L0-10 & Real-money transactions & F & Supplies real-world social or payment accounts, or explicitly redirects participants to transfers for virtual-item or account trading. \\
L0-11 & Cheating and boosting & F & Recommends, sells, or suggests cheats, scripts, or assistance software, or publicly promotes boosting services. \\
L0-12 & Personal privacy & M & Requests or supplies real phone numbers, home addresses, identity information, or similar private data. \\
L0-13 & Abuse and harassment & M & Uses family-directed abuse, regional discrimination, or sexual harassment; ordinary competitive frustration is permitted. \\
L0-14 & Automated advertising & m & Produces meaningless high-frequency repetition or unrelated commercial keywords. \\
\bottomrule
\end{tabularx}
\end{table}

\FloatBarrier

\begingroup
\footnotesize
\setlength{\tabcolsep}{4pt}
\renewcommand{\arraystretch}{1.0}
\setlength{\LTcapwidth}{\linewidth}
\begin{longtable}{@{}>{\raggedright\arraybackslash}p{0.045\linewidth}>{\raggedright\arraybackslash}p{0.19\linewidth}>{\raggedleft\arraybackslash}p{0.04\linewidth}p{\dimexpr0.725\linewidth-24pt\relax}@{}}
\caption{Game Interaction dimensions and checkbox criteria. Each numbered item corresponds to one binary checkbox. Weights are normalized within each scoring scope.}\label{tab:rubric_game}\\
\toprule
ID & Dimension & Wt. & Checkbox criteria \\
\midrule
\endfirsthead
\multicolumn{4}{@{}l}{\tablename~\thetable{} (continued)}\\
\toprule
ID & Dimension & Wt. & Checkbox criteria \\
\midrule
\endhead
\midrule
\multicolumn{4}{r@{}}{\textit{Continued on the next page}}\\
\endfoot
\bottomrule
\endlastfoot
\multicolumn{4}{@{}l}{\textit{Per-turn}} \\*
D1 & Game-chat naturalness & 6.5 & (1) Use casual, concise language. (2) Use particles naturally when present; terse replies without particles also pass. (3) Avoid formal written connectors. \\
D2 & Information density & 4.5 & (1) Match length to the relationship, situation, and emotion. (2) Avoid over-explaining. \\
D3 & Persona expression & 7 & (1) Match language to the persona's age. (2) Maintain personality traits. (3) Use catchphrases naturally when present; their absence also passes. \\
D4 & Rhythm and segmentation & 4 & (1) Use an appropriate number of message segments. (2) Keep each message independently readable. \\
D5 & Knowledge boundaries & 4 & (1) Avoid expertise beyond the persona. (2) Express uncertainty when a knowledge gap arises; no such gap also passes. \\
D6 & Game terminology & 10 & (1) Use class, equipment, and skill abbreviations correctly. (2) Use competitive shorthand naturally. (3) Avoid redundant explanations of familiar game terms. \\
D7 & Coordination and state awareness & 7.5 & (1) Reflect waiting, busyness, or multitasking when relevant. (2) Keep team or trade instructions clear and brief when issued. (3) Respond appropriately to reminders or waiting when present. \\
D8 & Competitive tone & 7.5 & (1) Use natural affective tone; symbols or particles are optional. (2) Express exploration or uncertainty in bargaining contexts. (3) Match tone to the relationship, such as caution with strangers and directness with friends. \\
D9 & Virtual-item descriptions & 5 & (1) Ground item or account values in the game market when discussed. (2) Use in-game units for quantities and currency when relevant. (3) Distinguish game currency from real money, except in scenarios explicitly involving prohibited trading. \\
D10 & Fragmented expression & 5 & (1) Use short utterances, allowing omitted subjects or predicates. (2) Allow harmless typos and abbreviations without requiring them; avoid artificially formal prose. (3) Avoid overly complete, formal written constructions. \\
D11 & Unpunctuated chat style & 50 & (1) Do not connect independent phrases within a message using formal punctuation such as commas, colons, dashes, or parentheses; sentence-final punctuation and one-character replies are allowed. (2) Do not pack independent phrases into one message using spaces or separators; express one point per message, including in single-message turns. \\
\midrule
\multicolumn{4}{@{}l}{\textit{Holistic}} \\*
E1 & Context consistency & 7 & (1) Avoid asking again for known information. (2) Keep facts consistent. \\
E2 & Scenario and relationship fit & 5 & (1) Respond to the triggering event. (2) Respect relationship boundaries. \\
E3 & Social-emotional fit & 5.5 & (1) Respond to explicitly strong emotion; natural teasing or complaints can qualify. (2) Do not ignore an explicit request for comfort; direct advice or complaints among friends can pass. \\
E4 & Initiative & 4.5 & (1) Advance the exchange at least once through follow-up, uptake, or a new topic; questions are not required every turn. (2) Express personal experience, a complaint, or a stance at least once. \\
E5 & Dialogue participation & 2 & (1) Ensure both participants contribute substantively. \\
E6 & Goal completion & 7.5 & (1) Have a clear in-game motive. (2) Exchange the information needed for that motive; a complete checklist of price, time, and ID is not required. \\
E7 & Relationship boundaries & 7.5 & (1) Maintain caution in stranger-to-stranger trading when applicable. (2) Use appropriately casual and direct interaction among friends or guild members. (3) Avoid abrupt, unsupported shifts in familiarity. \\
E8 & Fragmentation and immediacy & 25 & (1) Keep most evaluated-role messages to one information point, splitting independent phrases across messages. (2) Maintain an instant-messaging style rather than structured paragraphs. (3) Vary message length, including very short and longer contributions. \\
E9 & Conversational asymmetry & 15 & (1) Allow natural differences in contribution volume or frequency; balanced interaction also passes when natural. (2) Allow consecutive messages from one speaker; natural one-to-one alternation also passes. (3) Allow abrupt topic shifts, endings, or logical jumps; coherent topic development also passes when natural. \\
\end{longtable}
\endgroup

\FloatBarrier

\begin{table}[!htbp]
\centering
\footnotesize
\setlength{\tabcolsep}{4pt}
\renewcommand{\arraystretch}{1.0}
\caption{L0 validity constraints for Long-Horizon Character. F denotes fatal severity. Every listed violation invalidates the evaluated-role case.}
\label{tab:l0_character}
\begin{tabularx}{\linewidth}{@{}l>{\raggedright\arraybackslash}p{0.19\linewidth}cX@{}}
\toprule
ID & Constraint & Sev. & Operational evidence \\
\midrule
L0-01 & AI identity exposure & F & Discloses being an AI or language model. \\
L0-02 & Safety/ethics & F & Gives clearly illegal, dangerous, or harmful advice. \\
L0-03 & Hard repetition & F & Exactly copies a historical message. \\
L0-04 & Output contract & F & Missing or malformed JSON, role, response, content, or timestamp fields. \\
L0-05 & Persona facts & F & Contradicts explicit gender, age, occupation, relationship, or location. \\
\bottomrule
\end{tabularx}
\end{table}

\FloatBarrier

\begingroup
\footnotesize
\setlength{\tabcolsep}{4pt}
\renewcommand{\arraystretch}{1.0}
\setlength{\LTcapwidth}{\linewidth}
\begin{longtable}{@{}>{\raggedright\arraybackslash}p{0.045\linewidth}>{\raggedright\arraybackslash}p{0.19\linewidth}>{\raggedleft\arraybackslash}p{0.04\linewidth}p{\dimexpr0.725\linewidth-24pt\relax}@{}}
\caption{Long-Horizon Character dimensions and checkbox criteria. Each numbered item corresponds to one binary checkbox. Weights are normalized within each scoring scope.}\label{tab:rubric_character}\\
\toprule
ID & Dimension & Wt. & Checkbox criteria \\
\midrule
\endfirsthead
\multicolumn{4}{@{}l}{\tablename~\thetable{} (continued)}\\
\toprule
ID & Dimension & Wt. & Checkbox criteria \\
\midrule
\endhead
\midrule
\multicolumn{4}{r@{}}{\textit{Continued on the next page}}\\
\endfoot
\bottomrule
\endlastfoot
\multicolumn{4}{@{}l}{\textit{Per-turn}} \\*
D1 & Interpretable expression & 10 & (1) Make intent, attitude, or reference recoverable from the message and local context; fragments, typos, interruptions, and short acknowledgments may pass. (2) Keep text usable as dialogue; formality, length, and incomplete sentences alone are not failures. \\
D2 & Autonomous responses & 10 & (1) Fail only after three consecutive replies neither engage the partner nor provide natural listening signals; brief acknowledgments during narration pass. (2) Fail only after three consecutive substantive replies echo or paraphrase the partner without adding information or stance; ordinary restatement and listening signals pass. \\
D3 & Appropriate interaction wording & 10 & (1) Fail only when three consecutive replies keep pushing questions, proposals, or plans after an answer, disengagement, or topic departure. (2) Fail when three consecutive visible replies repeat the fixed agree/restate--elaborate--question/plan structure. (3) Respect an explicit refusal, stop request, or topic boundary immediately; no explicit boundary passes. \\
D4 & Grounded emotional wording & 10 & (1) Do not interpret emotion as the opposite of the latest explicit cue; emotional expression is optional. (2) Fail only when three consecutive visible replies exactly mirror emotional intensity without independent response or change; natural resonance passes. \\
D5 & Local template avoidance & 10 & (1) Fail when three substantive replies preserve a near-identical functional sequence or syntactic frame across different inputs; isolated similarity and listening signals pass. (2) Fail when three consecutive substantive replies strongly overlap without new information or stance; one repetition, two retransmissions, and listening signals may pass. \\
\midrule
\multicolumn{4}{@{}l}{\textit{Holistic}} \\*
E1 & Context consistency & 8 & (1) Avoid re-asking explicitly answered information; clarification and unanswered questions pass. (2) Avoid direct, unconditional contradiction of an established fact; missing persona details or qualified disclosures are not sufficient evidence of failure. (3) Attribute statements, experiences, opinions, and actions to the correct speaker. (4) Track confirmed decisions and states. (5) Avoid treating invented or stale memories as current shared facts or creating unexplained conflicts with established information. (6) Do not fabricate shared participation or assert unprovided partner facts as common knowledge; new personal disclosures are allowed. \\
E2 & Response specificity & 15 & (1) Adapt substantive content to the input; fail for interchangeable generic replies that ignore distinct inputs, not merely similar rhetorical order. (2) Avoid sustained restatement or repair without new facts, stance, or necessary clarification; listening signals are excluded. (3) Adapt the response's semantic function to the interaction; length, register, message count, and continuation are not criteria, and repeated functions may pass when relevant. \\
E3 & Emotional adaptation & 7 & (1) Avoid persistent copying of emotional polarity or intensity across different interactions without independent response or change; natural resonance passes. (2) Avoid repeated unsupported attributions of the partner's emotions; no attribution or one reasonable guess may pass. (3) Do not persistently override explicit discomfort, refusal, or boundaries with incompatible enthusiasm; coolness or limited elaboration alone is not penalized. \\
E4 & Relevant initiative & 10 & (1) Do not repeatedly replace a response to explicit current content with unrelated questions, plans, invitations, or topic reopenings; relevant associations and continuation pass. (2) Do not persistently recast explicit information, questions, or boundaries as unrelated suggestions or inquiries; initiative is optional and listening passes. (3) Do not assert unprovided partner difficulties, schedules, motives, abilities, or goals as facts to justify initiative; new topics, personal disclosures, and clearly marked guesses awaiting confirmation may pass. \\
E5 & Scenario and factual grounding & 10 & (1) Address explicit current questions, requests, boundaries, or important facts rather than persistently replacing them with a generic script. (2) Do not assert absent or contradicted scenario premises, shared experiences, or unprovided partner circumstances as facts; personal disclosures, explicit hypotheses, and examples may pass. (3) Respect explicit relationship, address, or contact boundaries; changes in warmth, directness, or temporary coolness alone may pass. \\
\end{longtable}
\endgroup

\FloatBarrier

\subsection{Scoring, Acceptance, and Diagnostics}
\label{app:scoring}

For dimension $k$ with $m_k$ binary criteria, its score is the fraction of passed checkboxes:
\begin{equation}
q_k=\frac{1}{m_k}\sum_{j=1}^{m_k}\mathbb{I}[b_{k,j}=1].
\label{eq:app_checkbox_score}
\end{equation}
Let $T_r$ be the set of generated messages from evaluated role $r$, and let $w_k$ be the fixed rubric weight of dimension $k$. The per-turn and holistic scores normalize these weights within their respective dimension sets:
\begin{equation}
\begin{aligned}
S_{\mathrm{turn}}&=\frac{1}{|T_r|}\sum_{t\in T_r}
\frac{\sum_{k\in\mathcal{D}_{\mathrm{turn}}}w_kq_{t,k}}
{\sum_{k\in\mathcal{D}_{\mathrm{turn}}}w_k},\\
S_{\mathrm{dialogue}}&=
\frac{\sum_{k\in\mathcal{D}_{\mathrm{dialogue}}}w_kq_k}
{\sum_{k\in\mathcal{D}_{\mathrm{dialogue}}}w_k}.
\end{aligned}
\label{eq:app_scope_scores}
\end{equation}
For validity indicator $v\in\{0,1\}$, the overall Score combines the two scopes with equal weight:
\begin{equation}
S=v\bigl(0.5S_{\mathrm{turn}}+0.5S_{\mathrm{dialogue}}\bigr).
\label{eq:app_overall_score}
\end{equation}
Score is reported on the $[0,1]$ scale; the equivalent 100-point value retained in evaluation traces is $100S$. These evaluation weights remain fixed across models and checkpoints. DiAPO's adaptive reward weights affect training but do not redefine benchmark scores.

The reported ACC@$T$ is a scenario-level thresholded pass rate. For $N$ evaluated bindings, let $v_{n,r}$ and $S_{n,r}$ denote the validity indicator and final score for role $r$ in binding $n$. The pass rate is computed as:
\begin{equation}
\operatorname{ACC}@T=\frac{1}{N}\sum_{n=1}^{N}
\prod_{r\in\{a,b\}}\mathbb{I}[v_{n,r}=1\ \land\ S_{n,r}\geq T],
\qquad T\in\{0.85,0.90,0.95\}.
\label{eq:app_threshold_accuracy}
\end{equation}
A binding passes only when both evaluated roles are valid and meet the threshold. The evaluation reports also retain a stricter all-checkbox-pass diagnostic: an evaluated-role case passes only if it is L0-valid and every checkbox passes in all its evaluated turns and its holistic scorecard. This diagnostic is aggregated over valid cases, with valid rate and L0 failure counts recorded separately over all scheduled cases. It is distinct from the scenario-level ACC@$T$ reported in the result tables.

Detailed evaluation traces retain failed criteria and judge rationales, alongside per-turn, holistic, per-dimension, and category-level summaries. These records support capability analysis and data selection, including rejection, repair, preference-pair construction, and regression testing.

\subsection{Interaction Protocol Details}
\label{app:sati}

MindFlow maintains a separate private draft for each participant on a shared virtual timeline. At scheduling event $n$, participant $i$ observes:
\begin{equation}
 x_{i,n}=[p_i,p_j,s,m_{i\rightarrow j},C_{i,n},
 \operatorname{tail}_N(h_n),b_{i,n-1},\tau_n].
\label{eq:app_interaction_state}
\end{equation}
Here, $p_i$ and $p_j$ are the persona cards, $s$ is the scenario card, $m_{i\rightarrow j}$ contains partner-specific long-term memory, $C_{i,n}$ summarizes older context, and $\operatorname{tail}_N(h_n)$ contains the $N$ most recent messages. The state also includes any unsent draft $b_{i,n-1}$ and the current virtual time $\tau_n$. Persona cards describe identity, knowledge boundaries, and expression habits; scenario cards specify the relationship, triggering events, and participant-specific circumstances. Partner memory remains distinct from live history so that past experience is not confused with a current event.

Each participant keeps at most one pending draft. A proposed delay $\delta_{i,n}$ determines its absolute delivery time $a_{i,n}$, and the scheduler advances to the earliest pending delivery:
\begin{equation}
\begin{aligned}
a_{i,n}&=\tau_n+\operatorname{clip}(\delta_{i,n},0,\Delta_{\max}),\\
\tau_{n+1}&=\min_{i:\,b_{i,n}\ne\emptyset}a_{i,n},\qquad
\mathcal{I}^{\mathrm{send}}_n=\{i:b_{i,n}\ne\emptyset,\ a_{i,n}=\tau_{n+1}\}.
\end{aligned}
\label{eq:app_delivery_scheduler}
\end{equation}
The bound $\Delta_{\max}$ limits the proposed delay. All messages in $\mathcal{I}^{\mathrm{send}}_n$ are delivered together, preserving simultaneous delivery for equal timestamps as in Section~\ref{sec:mindflow}. Both participants then observe the updated history and update their buffers. A participant may retain, revise, postpone, cancel, or replace an unsent draft; the sender may also prepare a follow-up without waiting for a reply. Withdrawal clears the pending draft but permits reactivation when the partner speaks. The interaction ends when both buffers are empty or a runtime limit is reached.

Recent dialogue and older memory remain distinguishable, while the complete trajectory is retained for CAPS-Eval. Older context can summarize facts, emotional changes, commitments, and unresolved topics, and partner memory can be updated between sessions. Training synthesis additionally constrains interaction length to reduce premature closure; these synthesis controls are distinct from the autonomous termination behavior evaluated by the benchmark.

\subsection{Benchmark Workflow}
\label{app:workflow}

Table~\ref{tab:workflow} summarizes case construction, dialogue generation, validity checking, scoring, and aggregation. Each binding is evaluated from both participant roles under the same scheduling conditions. The dimension identifiers shown in the table refer to Everyday Chat; the other subsets follow the same workflow with their task-specific rubrics.

\begin{table}[!htbp]
\centering
\small
\caption{Benchmark execution workflow. The evaluated role is the persona controlled by the tested model in the current case.}
\label{tab:workflow}
\begin{tabularx}{\linewidth}{@{}lX@{}}
\toprule
Stage & Operation \\
\midrule
Case construction & Load a binding $(p_a,p_b,s)$ and create two cases, one with the tested model as $p_a$ and one with it as $p_b$. \\
Dialogue generation & Instantiate both personas with the same scenario, opener, virtual start time, duration horizon, and single-draft scheduler; save dialogue and raw model-call traces. \\
L0 gate & Check deterministic identity, medium, co-presence, structure, and repetition failures; mark fatal cases invalid before soft scoring. \\
Per-turn scoring & For each evaluated-role turn, call the judge with the target turn and up to $W_{\mathrm{eval}}$ preceding turns; compute D1--D5 checkbox scores. \\
Holistic scoring & Call the judge once with the full dialogue and evaluated role marker; compute E1--E5 checkbox scores. \\
Aggregation & Aggregate Score and paired ACC@$T$; retain validity rates, case details, dimension and category breakdowns, and all-checkbox-pass diagnostics. \\
\bottomrule
\end{tabularx}
\end{table}

\FloatBarrier

\subsection{Judge Robustness}
\label{app:judge_robustness}

Table~\ref{tab:judge_robustness} compares Everyday Chat scores under human annotation and two automatic judges. Claude-4.6 Thinking ranks first under all three judges. When Qwen3.5-397B-A17B serves as the judge, Claude-4.6 Thinking and Gemini-3.5 Flash tie for first place. GPT-5.5 and Qwen3.5-9B rank fifth and sixth, respectively, under all three judges. The middle ranks vary: Qwen3.5-397B-A17B ranks third under human annotation but fourth under both automatic judges, and DeepSeek-V4-Flash ranks fourth under human annotation but third or tied second under automatic evaluation. Thus, the broad ordering is consistent, while absolute scores and some pairwise rankings remain judge-dependent.

\begin{table}[!htbp]
\centering
\centering
\footnotesize
\caption{Everyday Chat Score (rank among the compared models) under three judges.}
\label{tab:judge_robustness}
\setlength{\tabcolsep}{4pt}%
\begin{tabular}{@{}lccc@{}}
\toprule
\textbf{Model} & \textbf{Human Annotators} & \textbf{Qwen3.5-397B-A17B} & \textbf{DeepSeek-V4-Flash} \\
\midrule
Claude-4.6 Thinking & 1.0000 (1) & 0.9780 (1) & 0.9830 (1)  \\
Gemini-3.5 Flash & 0.9560 (2) & 0.9780 (1) & 0.9790 (2)  \\
Qwen3.5-397B-A17B & 0.9410 (3) & 0.9410 (4) & 0.9650 (4)  \\
DeepSeek-V4-Flash & 0.9340 (4) & 0.9680 (3) & 0.9790 (2)  \\
GPT-5.5 & 0.8900 (5) & 0.8820 (5) & 0.9140 (5)  \\
Qwen3.5-9B & 0.6260 (6) & 0.6410 (6) & 0.7390 (6)  \\
\bottomrule
\end{tabular}
\end{table}

\FloatBarrier

\FloatBarrier
\section{More Details of Anthropomorphism Learning}
\label{app:learning}

\subsection{Persona--Scenario Expansion and SFT Data}
\label{app:seeds}

SEEDS expands the conditions that generate training behavior. Repeated sampling from a small set of persona--scenario bindings can produce many utterances while leaving the underlying relationships and events largely unchanged. Persona--scenario expansion diversifies both the participating personas and their interaction contexts, including relationships, triggering events, and initial circumstances. A binding $(p_a,p_b,s)$ remains the basic unit, preserving compatibility between the participants and the situation.

Two generation routes address different coverage gaps. Scenario-conditioned expansion selects a branch of the fixed scenario taxonomy, creates a concrete triggering situation, and generates two personas suited to its relationship and events. Persona-conditioned expansion selects an underused existing pair and constructs a situation from their shared interests, differences, and plausible social relationship. Coverage counts guide the former toward missing or underrepresented branches, while binding histories guide the latter toward less-explored pairs. New leaf scenarios are generated within the catalog's existing categories.

Generated cards undergo checks for required fields and category membership before receiving identifiers and becoming eligible bindings. Persona and scenario quotas limit concentration on frequently reused characters or settings. Each binding also has a rejection budget. Once all eligible bindings reach their quotas or failure limits, the system creates new bindings and resumes synthesis. Scenario cards establish an interaction's starting conditions, including events and emotional context, without prescribing its conversational conclusion.

MindFlow generates candidate dialogues from these environments. Deterministic checks reject repetition and out-of-range lengths, after which CAPS-Eval assesses the trajectory. Accepted cases enter the data pool and update usage counts. Rejected cases retain their failure diagnostics and consume the binding's retry budget. Thus, feedback affects both which trajectories become training data and which environments remain eligible for further sampling. The dialogue synthesis configuration requires at least $T_{\min}^{\mathrm{syn}}$ turns and constrains early closure; unrestricted evaluation can expose a broader termination policy. Failure-driven rewriting of behavioral rules is an optional extension to this expansion-and-filtering loop.

Accepted trajectories supply recorded state--decision pairs. Repaired trajectories are included only after revalidation. Supervised training minimizes
\[
  \mathcal{L}_{\mathrm{SFT}}(\phi)=
  -\mathbb{E}_{(x,d)\sim\mathcal{D}_{\mathrm{SFT}}}
  \log\pi_\phi(d\mid x).
\]
Through SFT, the policy learns to generate dialogue content and schedule its delivery based on the current interaction context. Table~\ref{tab:sft_statistics} summarizes the training data for the three subsets. The synthesized Everyday Chat dialogue collection involves 290 persona cards and 184 scenario cards. Expanding the dialogue messages into individual training examples and cleaning them yields 7,774 SFT samples. For Game Interaction and Long-Horizon Character, card counts refer to the unique personas and scenarios represented in the prepared SFT files. Each SFT sample is an individual training record rather than an entire dialogue. Table~\ref{tab:everyday_chat_sft} further breaks down the Everyday Chat training data by scenario category.

\begin{table}[!htbp]
\centering
\small
\caption{SFT data statistics across the three subsets.}
\label{tab:sft_statistics}
\begin{tabular}{@{}lrrr@{}}
\toprule
Subset & Scenario cards & Persona cards & SFT samples \\
\midrule
Everyday Chat & 184 & 290 & 7,774 \\
Game Interaction & 18 & 36 & 430 \\
Long-Horizon Character & 178 & 314 & 22,645 \\
\bottomrule
\end{tabular}
\end{table}

\begin{table}[!htbp]
\centering
\small
\caption{Category distribution of synthesized SFT data for Everyday Chat.}
\label{tab:everyday_chat_sft}
\setlength{\tabcolsep}{4pt}
\begin{tabular}{@{}lrrrr@{}}
\toprule
Category & Scenario cards & Scenario share & SFT samples & Training share \\
\midrule
Interests & 38 & 20.65\% & 1,662 & 21.38\% \\
Daily chat & 15 & 8.15\% & 440 & 5.66\% \\
Emotional support & 21 & 11.41\% & 1,072 & 13.79\% \\
Technology & 15 & 8.15\% & 618 & 7.95\% \\
Family and parenting & 13 & 7.07\% & 491 & 6.32\% \\
Relationship development & 3 & 1.63\% & 116 & 1.49\% \\
Finance & 17 & 9.24\% & 909 & 11.69\% \\
Pets and family life & 17 & 9.24\% & 640 & 8.23\% \\
News & 10 & 5.43\% & 495 & 6.37\% \\
Gaming & 13 & 7.07\% & 486 & 6.25\% \\
Work communication & 22 & 11.96\% & 845 & 10.87\% \\
\midrule
Total & 184 & 100.00\% & 7,774 & 100.00\% \\
\bottomrule
\end{tabular}
\end{table}

\FloatBarrier

\subsection{DiAPO Implementation Details}
\label{app:diapo}

This section details the rollout observations, task--capability matching, capability updates, and policy optimization used by DiAPO in Section~\ref{sec:diapo}. We retain the notation $q\in\{\mathrm{per},\mathrm{whole}\}$ for evaluation granularity and add a batch index $b$ to quantities that evolve during training.

\paragraph{Rollout groups and capability observations.}
The task group $g$ in Section~\ref{sec:diapo} is a \emph{rollout group}: repeated trajectories generated with the same persona pair, scenario, and evaluated role. The two participant roles form separate groups, and a training batch may contain multiple rollout groups. To make aggregation within each trajectory explicit, let $u$ index an evaluated-role trajectory and $j$ one of its $T_u$ trainable dialogue decisions. The per-turn and holistic dimension scores are $s^{\mathrm{per}}_{u,j,k}$ and $s^{\mathrm{whole}}_{u,k}$, respectively, each given by its checkbox pass fraction. For Game Interaction, $|\mathcal{K}_{\mathrm{per}}|=11$ and $|\mathcal{K}_{\mathrm{whole}}|=9$. Trajectory-level observations and the mean performance of rollout group $g$ are computed as:
\begin{equation}
\begin{aligned}
  o^{\mathrm{per}}_{u,k}
  &=\frac{1}{T_u}\sum_{j=1}^{T_u}s^{\mathrm{per}}_{u,j,k},
  \qquad
  o^{\mathrm{whole}}_{u,k}=s^{\mathrm{whole}}_{u,k},\\
  \bar{s}^{q}_{g,k}
  &=\frac{1}{N_g}\sum_{u\in\mathcal{I}_g}o^{q}_{u,k},
  \qquad N_g=|\mathcal{I}_g|.
\end{aligned}
\label{eq:app_diapo_observations}
\end{equation}
Here, $\mathcal{I}_g$ contains the valid evaluated-role trajectories in group $g$. Averaging within each trajectory first prevents longer conversations from dominating capability observations. Each rollout--role pair is counted once, and fixed opening messages are excluded from the trainable decisions.

\paragraph{Capability diagnosis and task--capability matching.}
At the start of batch $b$, $\theta^q_{k,b}$ summarizes historical performance on dimension $k$ at granularity $q$. The estimates are initialized from dimension-wise capability priors or restored from a saved state. Following Equation~\ref{eq:diagnostic_coefficients}, the capability-deficit coefficient and matching center are computed as:
\begin{equation}
  c^q_{k,b}=\epsilon+(1-\theta^q_{k,b})^p,
  \qquad
  m^q_{k,b}=\operatorname{clip}(\theta^q_{k,b}-\delta,0,1).
\label{eq:app_diapo_deficit}
\end{equation}
Lower historical performance gives a larger deficit coefficient, while $\epsilon>0$ preserves a learning signal for stronger dimensions. DiAPO combines this diagnosis with task--capability matching inspired by the Zone of Proximal Development \citep{Vygotsky1978Mind}. The matching factor uses the mean performance of each rollout group to assess whether the corresponding dimension falls within an effective learning range:
\begin{equation}
\begin{aligned}
  \omega^q_{g,k,b}
  &=\operatorname{clip}\!\left(
  \exp\!\left[-\frac{(\bar{s}^q_{g,k}-m^q_{k,b})^2a^q_{g,k,b}}{2\sigma^2}\right],
  \omega_{\min},1\right),\\
  a^q_{g,k,b}
  &=\begin{cases}
    \gamma_{\mathrm{easy}},&\bar{s}^q_{g,k}>\theta^q_{k,b},\\
    \gamma_{\mathrm{hard}},&\bar{s}^q_{g,k}<m^q_{k,b},\\
    1,&\text{otherwise}.
  \end{cases}
\end{aligned}
\label{eq:app_diapo_matching}
\end{equation}
The matching factor is largest when the group's mean score is close to the matching center. The score provides an empirical performance signal rather than an independent measure of task difficulty. The asymmetric coefficients control attenuation on the easy and hard sides of the learning range. All trajectories within the same rollout group use the same matching factor and dimension weights, so alternative rollouts generated under identical conditions are compared using a shared reward weighting. These weights combine rubric importance, capability deficit, and task--capability matching as in Equation~\ref{eq:adaptive_reward}.

\paragraph{Capability updates.}
Rewards for batch $b$ use capability estimates $\theta^q_{k,b}$ frozen at the start of that batch. After reward computation, the estimates are updated from raw dimension scores of valid trajectories. Let $\bar{o}^q_{k,b}$ and $v^q_{k,b}$ denote the mean and sample variance of the $N^q_{k,b}$ valid trajectory-level observations. The scalar Kalman update \citep{Kalman1960Filtering} takes the following form:
\begin{equation}
\begin{aligned}
  V^q_{k,b}&=R_{\min}+\frac{v^q_{k,b}}{N^q_{k,b}},
  &P^{q,-}_{k,b}&=P^q_{k,b}+Q,\\
  K^q_{k,b}&=\frac{P^{q,-}_{k,b}}{P^{q,-}_{k,b}+V^q_{k,b}},
  &P^q_{k,b+1}&=(1-K^q_{k,b})P^{q,-}_{k,b},\\
  \theta^q_{k,b+1}
  &=\operatorname{clip}\!\left(
    \theta^q_{k,b}+K^q_{k,b}(\bar{o}^q_{k,b}-\theta^q_{k,b}),0,1\right).
\end{aligned}
\label{eq:app_diapo_capability_update}
\end{equation}
Here, $P^q_{k,b}$ represents uncertainty in the capability estimate, $Q$ is process noise, and $R_{\min}$ is the observation-noise floor. For a single observation, $v^q_{k,b}=0$. Observation variance controls how strongly a noisy batch revises the prior. Updating from raw dimension scores keeps capability estimation separate from adaptive reward weighting.

\paragraph{Policy optimization.}
DiAPO converts the mixed rewards in Equation~\ref{eq:final_reward} into group-relative advantages and optimizes a clipped policy surrogate, following \citet{Shao2024DeepSeekMath}. Let $\widehat{A}_{u,j}$ denote the advantage for decision $j$ in trajectory $u$, and let $r_{u,j,\ell}(\phi)$ be the current-to-rollout policy probability ratio for generated token $\ell$. The token-level surrogate is defined as:
\begin{equation}
  \mathcal{J}_{u,j,\ell}(\phi)=\min\!\left[
    r_{u,j,\ell}(\phi)\widehat{A}_{u,j},
    \operatorname{clip}\!\left(r_{u,j,\ell}(\phi),
    1-\varepsilon_{\mathrm{low}},1+\varepsilon_{\mathrm{high}}\right)
    \widehat{A}_{u,j}\right].
\label{eq:app_diapo_policy_surrogate}
\end{equation}
The clipping tolerances $\varepsilon_{\mathrm{low}}$ and $\varepsilon_{\mathrm{high}}$ bound changes in the policy probability ratio. The surrogate is aggregated over trained response tokens for policy optimization.

Each DiAPO iteration collects MindFlow rollouts, scores them with the fixed CAPS-Eval rubrics, and computes training rewards using the capability estimates available at the start of the batch. These rewards guide the policy update, while the raw dimension scores update the capability estimates used in subsequent batches. This procedure adapts the training objective to prioritize underdeveloped yet learnable capabilities; benchmark evaluation continues to use the same predefined rubrics and aggregation weights across checkpoints.

\clearpage
\section{More Details of Experimental Results}
\label{app:experimental_results}

\subsection{Baseline Comparison}
\label{app:baseline_comparison}

We compare the SEEDS--DiAPO pipeline against three dialogue methods that target human-like conversation across the three benchmark subsets.

\noindent \textbf{Methods.}
ECP \citep{luo-laban-2026-spasm} is the Egocentric Context Projection component of SPASM, which stores dialogue history in a perspective-agnostic representation and deterministically projects it into each agent's egocentric view before generation, thereby mitigating persona drift in long-horizon conversations; we evaluate ECP with the same Qwen3.5-9B backbone as our base model.
HumanLM \citep{wu2026humanlm} simulates users by aligning generated behavior with the user's evolving internal state rather than imitating surface-level responses.
OSIM \citep{zhou2026odyssim} is a foundation model for human behavior simulation.
All three are evaluated under the same Qwen3.5-397B-A17B judge and the same scenarios as our pipeline.

\noindent \textbf{Results.}
Table~\ref{tab:training_full} shows consistent gains from SEEDS-based SFT and DiAPO across the three benchmark subsets. SFT already surpasses ECP, HumanLM, and OSIM in overall Score on every subset, including ECP with the same Qwen3.5-9B backbone. The largest gap appears on Long-Horizon Character, where the strongest dialogue baseline reaches a Score of $0.1880$, compared with $0.7280$ after SFT and $0.9930$ after DiAPO.

DiAPO further improves upon SFT and achieves the highest overall Score and ACC@95 on each subset. Its ACC@95 reaches $0.7600$, $0.9200$, and $0.9661$ on Everyday Chat, Long-Horizon Character, and Game Interaction, respectively, compared with GRPO's $0.7200$, $0.7400$, and $0.8814$. These results show that anthropomorphism learning improves both average interaction quality and the proportion of scenarios in which both participants meet stringent evaluation criteria.

\begin{table}[t]
\centering
\small
\caption{Effectiveness of anthropomorphism learning across the three CAPS-Eval scenarios. The top block lists baseline dialogue methods (ECP \citep{luo-laban-2026-spasm}, HumanLM \citep{wu2026humanlm}, and OSIM \citep{zhou2026odyssim}); the bottom block traces the SEEDS--DiAPO pipeline on Qwen3.5-9B. SFT selects the best checkpoint on validation Score; RL rows report the best checkpoint of each method on each scenario. For the Game Interaction SFT setting, threshold pass rates use the 89/90/95 evaluation available for that sweep.}
\label{tab:training_full}
\setlength{\tabcolsep}{4pt}%
\resizebox{\linewidth}{!}{%
\begin{tabular}{@{}l|ccc|ccc|ccc@{}}
\toprule
& \multicolumn{3}{c|}{\textbf{Everyday Chat}} & \multicolumn{3}{c|}{\textbf{Long-Horizon Character}} & \multicolumn{3}{c}{\textbf{Game Interaction}} \\
\cmidrule(lr){2-4} \cmidrule(lr){5-7} \cmidrule(lr){8-10}
\textbf{Setting} & Score & Per-Turn & Holistic & Score & Per-Turn & Holistic & Score & Per-Turn & Holistic \\
\midrule
\multicolumn{10}{@{}l}{\emph{Baseline dialogue methods}} \\
ECP & 0.7140 & 0.8104 & 0.6178 & 0.0880 & 0.1335 & 0.0431 & 0.7310 & 0.6798 & 0.7822 \\
HumanLM & 0.6070 & 0.6405 & 0.5736 & 0.1880 & 0.2684 & 0.1079 & 0.7700 & 0.7701 & 0.7692 \\
OSIM & 0.3040 & 0.2035 & 0.4043 & 0.0720 & 0.1025 & 0.0407 & 0.3570 & 0.4665 & 0.2482 \\
\midrule
\multicolumn{10}{@{}l}{\emph{SEEDS--DiAPO pipeline (ours)}} \\
Base & 0.6410 & 0.7261 & 0.5558 & 0.2120 & 0.3374 & 0.0871 & 0.6620 & 0.5914 & 0.7320 \\
+ SFT & 0.9210 & 0.9677 & 0.8737 & 0.7280 & 0.9232 & 0.5338 & 0.8890 & 0.8348 & 0.9429 \\
+ SFT + RL (DiAPO) & 0.9801 & 0.9966 & 0.9636 & 0.9930 & 0.9948 & 0.9904 & 0.9926 & 0.9981 & 0.9871 \\
+ SFT + RL (GRPO) & 0.9731 & 0.9962 & 0.9501 & 0.9820 & 0.9978 & 0.9664 & 0.9879 & 0.9933 & 0.9825 \\
\midrule
& \multicolumn{3}{c|}{\textbf{Everyday Chat}} & \multicolumn{3}{c|}{\textbf{Long-Horizon Character}} & \multicolumn{3}{c}{\textbf{Game Interaction}} \\
\cmidrule(lr){2-4} \cmidrule(lr){5-7} \cmidrule(lr){8-10}
\textbf{Setting} & ACC@85 & ACC@90 & ACC@95 & ACC@85 & ACC@90 & ACC@95 & ACC@85 & ACC@90 & ACC@95 \\
\midrule
\multicolumn{10}{@{}l}{\emph{Baseline dialogue methods}} \\
ECP & 0.1200 & 0.0600 & 0.0200 & 0.0000 & 0.0000 & 0.0000 & 0.1017 & 0.0508 & 0.0169 \\
HumanLM & 0.0200 & 0.0000 & 0.0000 & 0.0000 & 0.0000 & 0.0000 & 0.4068 & 0.2373 & 0.1186 \\
OSIM & 0.0000 & 0.0000 & 0.0000 & 0.0000 & 0.0000 & 0.0000 & 0.0000 & 0.0000 & 0.0000 \\
\midrule
\multicolumn{10}{@{}l}{\emph{SEEDS--DiAPO pipeline (ours)}} \\
Base & 0.1200 & 0.0400 & 0.0000 & 0.0000 & 0.0000 & 0.0000 & 0.0000 & 0.0000 & 0.0000 \\
+ SFT & 0.8000 & 0.5200 & 0.2600 & 0.2000 & 0.0600 & 0.0600 & 0.2712 & 0.2373 & 0.0169 \\
+ SFT + RL (DiAPO) & 1.0000 & 0.9800 & 0.7600 & 1.0000 & 0.9600 & 0.9200 & 1.0000 & 1.0000 & 0.9661 \\
+ SFT + RL (GRPO) & 0.9800 & 0.8800 & 0.7200 & 1.0000 & 0.9000 & 0.7400 & 1.0000 & 1.0000 & 0.8814 \\
\bottomrule
\end{tabular}%
}
\end{table}
\FloatBarrier
\clearpage

\subsection{Dimension-Level Training Dynamics}
\label{app:diapo_dynamics}

\begin{figure}[!htbp]
  \centering
  \includegraphics[width=\linewidth]{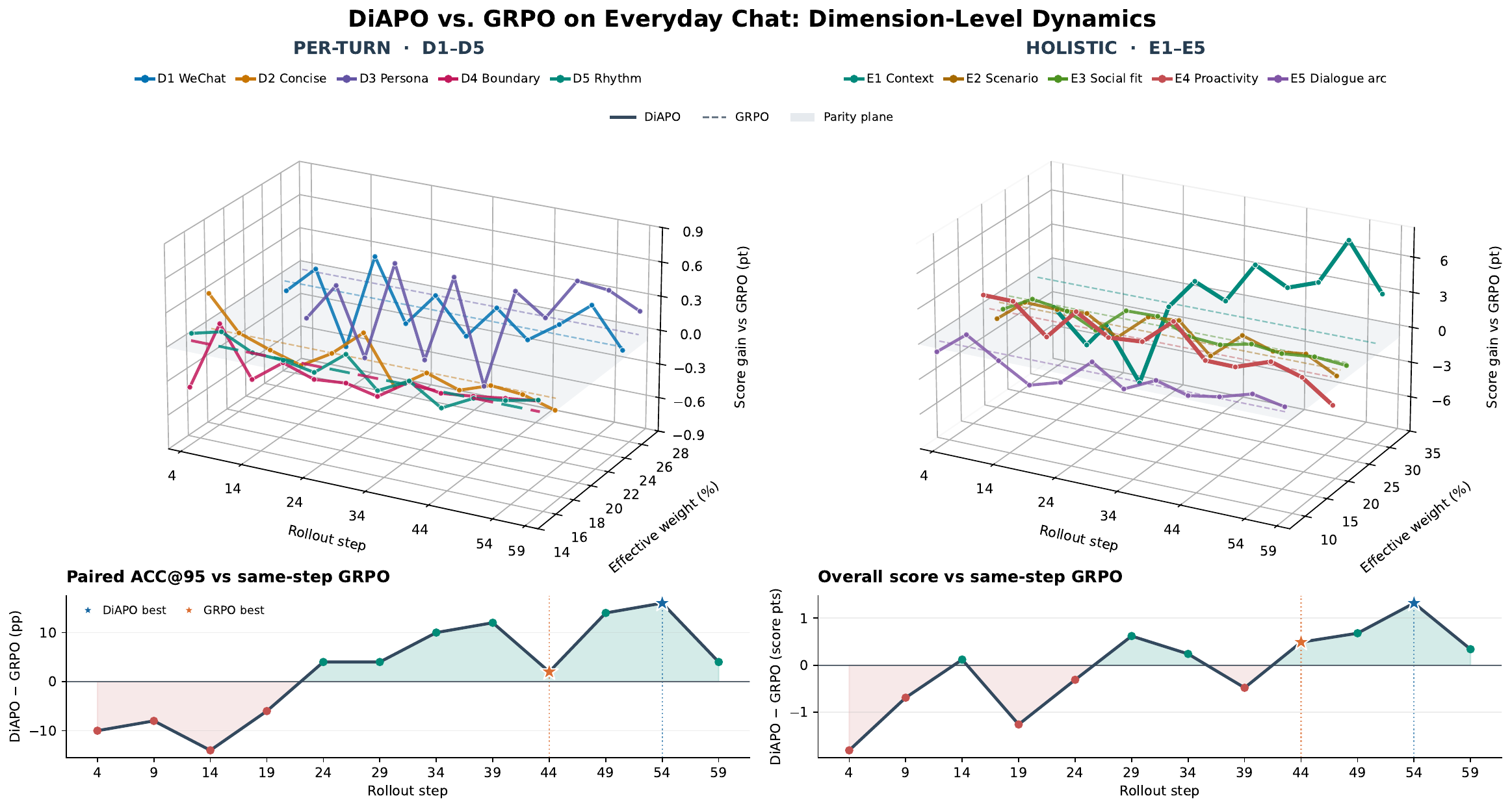}
  \caption{Dimension-level training dynamics and aggregate gains of DiAPO over GRPO on Everyday Chat. Top: trajectories for per-turn dimensions D1--D5 (left) and holistic dimensions E1--E5 (right). The $x$-, $y$-, and $z$-axes denote rollout step, effective dimension weight, and score gain over the same-step GRPO checkpoint, respectively; the translucent plane marks zero gain. Bottom: same-step DiAPO-minus-GRPO differences in paired ACC@95 (percentage points, left) and overall Score (points, right). Stars mark the best checkpoints of GRPO at step $44$ and DiAPO at step $54$.}
  \label{fig:rl_curves_full}
\end{figure}

Figure~\ref{fig:rl_curves_full} presents dimension-level training dynamics and aggregate performance comparisons between DiAPO and GRPO on Everyday Chat, complementing the main-text Figure~\ref{fig:rl_curves}.
The top panels show the trajectories of individual evaluation dimensions: per-turn D1--D5 on the left and holistic E1--E5 on the right. The $x$-axis denotes the rollout step, the $y$-axis denotes the effective weight assigned to each dimension by DiAPO, and the $z$-axis denotes its score gain over GRPO at the same training step. The translucent plane indicates zero gain, corresponding to parity with GRPO.

The bottom panels show DiAPO's aggregate advantage over GRPO at each matched training step, measured by paired ACC@95 on the left and overall Score on the right. Each point compares the DiAPO and GRPO checkpoints from the same step. The stars mark the best checkpoints of the two methods: step $44$ for GRPO and step $54$ for DiAPO.

The three-dimensional trajectories show that DiAPO dynamically adjusts the weights of different evaluation dimensions, with dimensions receiving larger weights tending to achieve greater gains over GRPO at later checkpoints. Consistent with these dimension-level improvements, the bottom panels show sustained improvement over GRPO during later training: the paired ACC@95 advantage becomes positive around step $24$ and remains positive thereafter, while both aggregate gains reach their largest plotted values at step $54$, DiAPO's best checkpoint. These results support DiAPO's core premise: capability diagnosis and task--capability matching help prioritize underdeveloped yet learnable dimensions, promoting balanced multidimensional improvement and stronger overall performance.

\end{document}